\ifdefined\XeTeXversion\else\pdfoutput=1\fi % arXiv: build with pdfLaTeX (skipped under XeTeX/tectonic previews)
\documentclass[10pt,twocolumn,letterpaper]{article}

\usepackage[applications,pagenumbers]{wacv} % CAMERA-READY version (applications track), with page numbers for arXiv
\usepackage{multirow}   % multi-row table cells (used in the supplementary)
\usepackage{makecell}

\providecommand{\cmark}{\checkmark}   % property present
\providecommand{\xmark}{\text{--}}    % property absent
\providecommand{\pmark}{$\circ$}      % partially present (footnoted)
\providecommand{\req}{\checkmark}     % input the method requires

\definecolor{wacvblue}{rgb}{0.21,0.49,0.74}
\usepackage[pagebackref,breaklinks,colorlinks,allcolors=wacvblue]{hyperref}

\def\wacvPaperID{*****} % *** Enter the WACV Paper ID here
\def\confName{WACV}
\def\confYear{2027}

\title{SpatialMem: Metric-Aligned Long-Horizon Video Memory for Language Grounding and QA}

\author{
Xinyi Zheng\textsuperscript{1,*} \quad
Yunze Liu\textsuperscript{2} \quad
Chi-Hao Wu\textsuperscript{2} \quad
Fan Zhang\textsuperscript{1} \quad
Hao Zheng\textsuperscript{1} \\
Wenqi Zhou\textsuperscript{1} \quad
Walterio W. Mayol-Cuevas\textsuperscript{1} \quad
Junxiao Shen\textsuperscript{1,2,\textdagger} \\[0.35em]
\textsuperscript{1}University of Bristol \qquad \textsuperscript{2}Memories.ai Research \\[0.25em]
{\tt\small \{wf24018, fan.zhang, me24271, bt24001, walterio.mayol-cuevas, junxiao.shen\}@bristol.ac.uk} \\
{\tt\small \{yunze.liu, chi-hao.wu, junxiao.shen\}@memories.ai} \\[0.25em]
{\small \textsuperscript{*}Equal contribution \quad \textsuperscript{\textdagger}Corresponding author}
}

\begin{document}
\maketitle

\begin{abstract}
SpatialMem is a memory-based system which integrates 3D geometry, semantics, and language with one another in one combined, queryable representation. Starting with casually recorded egocentric RGB video, SpatialMem recreates metrically scaled interior scenes, recognizes structural 3D anchors (walls, doors, windows) as the first-layer scaffold, and associates open-vocabulary object nodes with evidence patches, visual embeddings, and two-layer descriptions to 3D coordinates, enabling spatial-relation reasoning, language-directed retrieval, and QA without special sensors. We test at scale on public benchmarks. On VSI-Bench ($288$ videos, $5{,}130$ questions) SpatialMem scores $37.5$ overall with RGB alone, and its margin is precisely where a metric memory should earn it (room size $41.4$ vs.\ $24.4$ with a same-size frame model). On the episodic-memory task of OpenEQA ($953$ questions) the memory serialised to text is lower than the blind control, but as evidence-frame retrieval it recovers frame-level accuracy, matching uniform sampling at an equal eight-frame budget and pulling ahead as the budget tightens, with every answer grounded in linked 3D evidence. On $12$ ScanNet scenes evaluated using instance-level official annotations, the anchor-first hierarchy assigns the retrieved object to the proper structural parent where open-vocabulary 3D scene graphs essentially cannot ($0.24$ compared to $0.01$ with matched input), a capability difference instead of an overall win. The queryable index has a cost of $14$\,KB per scene without the linked evidence frames, and we make available the suites, memories and code.
\end{abstract}

\vspace{-0.5cm}
\section{Introduction}
\label{sec:intro}

Allowing autonomous agents like AR assistants or mobile robots to think about 3D indoor environments does not merely require frame-by-frame perception but rather a sustained and hierarchical spatial memory~\cite{blukis2022persistent}, which grounds language-based questions in a steady and metric world model by responding to relational questions such as ``what is on the table to the left of the window?'' with respect to persistent metric references, i.e., walls, doors, and windows, and thus object nodes are contextualized.

\begin{figure*}[t]
    \centering
    \includegraphics[width=\linewidth]{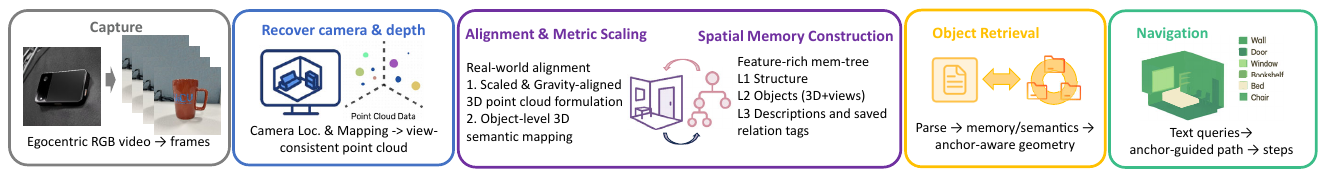}
    \caption{Schematic overview of the five-stage pipeline.}
    \label{fig:pipeline_schematic}
\end{figure*}

One of the challenges of implementing these systems is their reliance on depth sensors, IMUs, and calibrated Visual-SLAM~\cite{macario2022comprehensive}; to achieve low-cost implementation on phones and egocentric cameras, we construct a queryable 3D memory using casual egocentric RGB video only. Fig.~\ref{fig:pipeline_schematic} outlines five stages of the pipeline, starting with pose and depth recovery from egocentric RGB, moving through upright metric alignment and 3D lifting of structure and objects, to a rooted memory tree for path-based relational queries. Building such a unified memory from monocular 2D images raises four coupled problems. First, camera pose and depth must be recovered from an ill-posed monocular stream while remaining resistant to motion blur, occlusion, and light. The structure thus recovered should be translated into a shared metric frame whereby phrases such as ``next to the door'' or ``three meters behind the sofa'' have geometrical sense. Indoor scenes are also hierarchical because walls divide the rooms and rooms hold objects, and this context needs to be represented and preserved in order to reason compositionally and retrieve with multiple hops. Finally, once built, the memory must answer questions usefully despite noisy input so that it can be used interactively.

For these challenges, we propose a hierarchical 3D memory which incorporates the three domains of geometry, semantics, and language in one queryable structure to support metric relational queries as well as long-term object retrieval based on RGB video only. Every object has a two-layer description consisting of general features and then contextual relationships, and, extending monocular reconstruction~\cite{zou2022mononeuralfusion}, open-vocabulary segmentation~\cite{zhu2024survey}, and language models~\cite{zhang2025survey}, every response is based on 3D structural anchors with metric relations at low query latency.

We test SpatialMem with data at the benchmark level. On VSI-Bench~\cite{vsibench} ($5{,}130$ questions) its score of $37.5$ is higher than the frequency prior and GPT-4o, and comparable to a $7$B video model re-run using the same protocol. In the episodic-memory task of OpenEQA~\cite{openeqa} ($953$ questions), the way the memory is consumed is critical. When serialised to text, it performs worse than the blind language-prior control, like published caption-serialised scene graphs, but when used as \emph{retrieval}, it regains frame-level performance, equivalent to uniform sampling with the same frame budget, overtakes as the budget narrows, and is equivalent to a $72$B model with fifty frames while the answers stay grounded in connected 3D nodes and evidence. With $12$ ScanNet scenes evaluated based on official annotations, a privileged-input 3D scene graph localizes instances more accurately, whereas our anchor-first hierarchy attaches objects to structural parents where that representation cannot; this is a difference in capability, not an overall win; ablations attribute coverage to the object nodes and the hierarchy to the anchors.

Throughout the paper, \emph{guidance} means a set of steps to follow when retrieving information using a built memory as opposed to autonomous navigation, construction is done offline, scenes are static, and the quantitative statements are about relational QA and object retrieval. A \emph{spatial memory} is a structure that persists beyond the video and is re-used when answering queries (future work will involve incremental updates); \emph{anchors} are the structural elements of the first layer, with the \emph{object nodes} lying below them, an organization we have named an \emph{anchor-first hierarchy}; prior representations whose nodes are objects and which exclude background classes are referred to as \emph{3D scene graphs}.

We make several key contributions, namely (i) a hierarchical 3D memory that is purely based on egocentric RGB video, bringing geometry, semantics, and language into one queryable structure; (ii) two-layer object descriptions whose measured role is to bind attributes to nodes of the metric hierarchy rather than to improve retrieval; (iii) open-vocabulary querying through structural anchors having explicit metric relations grounded in 3D; (iv) a low-latency query system over a prebuilt memory; (v) public-benchmark scale analysis on VSI-Bench, the OpenEQA episodic-memory task, and a $12$-scene suite with official-annotation ground truth, compared to published persistent-memory systems, a privileged-input 3D scene graph and frame-based models, with component ablations of all four layers on two separate evaluations; (vi) a data release comprising the audited evaluation suite, a casually captured egocentric \emph{probe set} of $174$ human-verified episodes with their verification-only ground truth stated as such, and constructed memories for all $287$ public scenes.

\begin{figure*}[t]
    \centering
    \includegraphics[width=\linewidth]{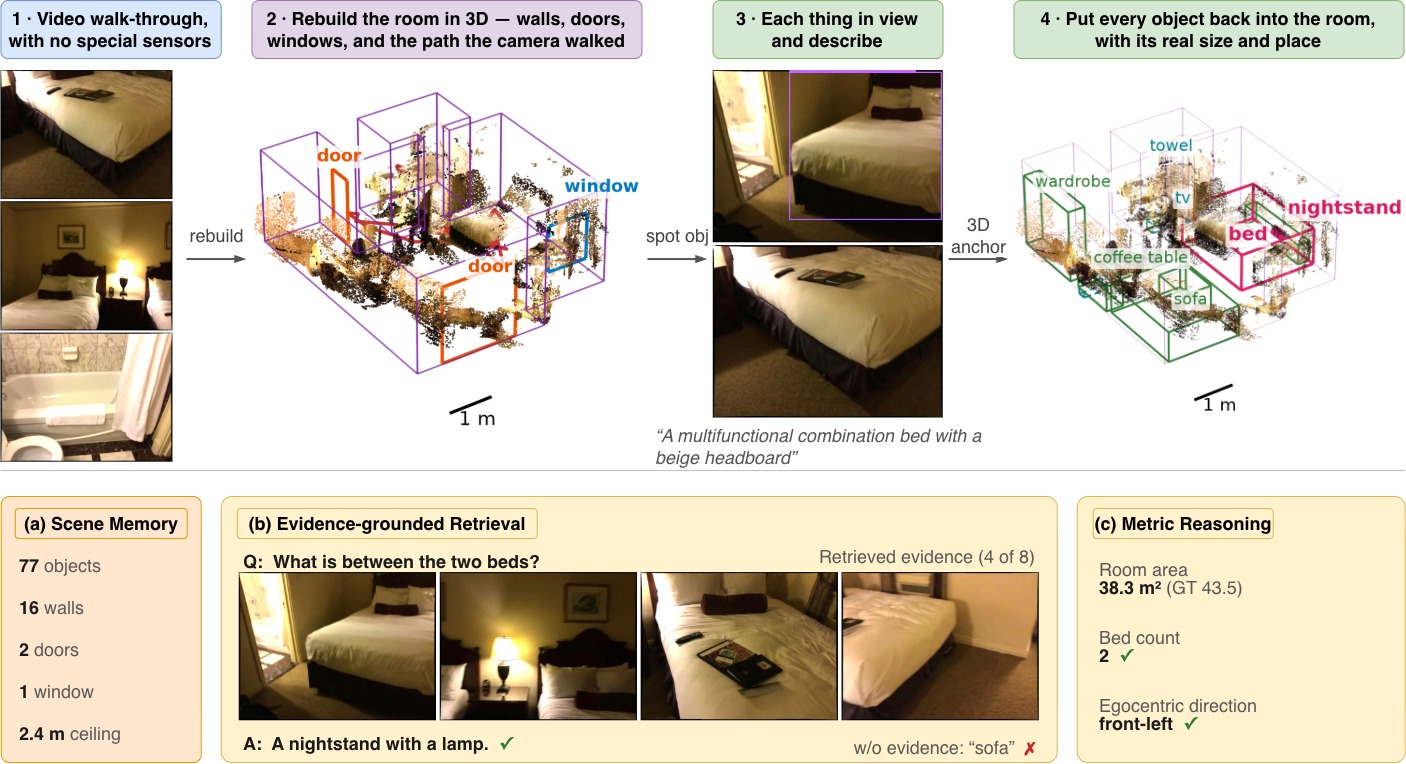}
    \caption{The pipeline applied to a real capture (\texttt{scene0645\_00}).}
    \label{fig:pipeline}
\end{figure*}

\section{Related Work}
\label{sec:related_work}
In our review of four areas we examine how they help in performing metric, relational and memory-based queries. Traditional 3D reconstruction tools like COLMAP~\cite{colmap} and ORB-SLAM2~\cite{mur2017orb} recover structure through feature correspondence and bundle adjustment, but are vulnerable to calibration on egocentric video with motion blur~\cite{mur2017orb, nerf}. Learning-based approaches estimate depth and pose in RGB, and transformer models, including VGGT~\cite{vggt} and SLAM3R~\cite{liu2025slam3r}, enhance speed and generalization. To us, these are exchangeable back ends as long as the recovered geometry is stable enough to support anchors and objects. For indoor layout understanding, approaches like RGB-D room boxes and wall/door/window detectors have structure but limited relational expression~\cite{zou2019manhattan,sceneParsing1,sceneParsing2}. Open-vocabulary tools (CLIP~\cite{clip}, GLIP~\cite{glip}, GroundingDINO~\cite{dino}, SAM~\cite{sam}, captioning~\cite{imgCaption}) increase category coverage but remain 2D and frame-local; scene graphs~\cite{hovsg} and LLM-assisted parsers~\cite{LLMparsing} add structure but lack a shared metric frame across views for determining direction, distance and visibility~\cite{hughes2022hydra}. The most similar systems to ours differ in \emph{what they store} and \emph{what inputs they require}; the capability matrix in Appendix~\ref{sec:supp-capability} separates the two, and one combination is empty. Methods that preserve a persistent metric 3D memory, such as open-vocabulary maps~\cite{openscene,openmask3d} or 3D scene graphs~\cite{hughes2022hydra,conceptgraphs,hovsg,graphpad}, use posed RGB-D. Methods operating on monocular RGB either lack persistent structure (frame-sampling VLMs), preserve a graph without metric geometry (VL-KnG~\cite{vlkng}), or integrate geometry into weights by tuning spatial instructions (VLM-3R~\cite{vlm3r}); 3D-Mem~\cite{threedmem} memorizes views as snapshots rather than a metric object map.

No prior system combines these two properties. \emph{Metric scale} allows a memory to respond in meters rather than merely ranking relations; the \emph{anchor-first hierarchy} determines what a location can be named relative to. ConceptGraphs does not include background classes, so walls cannot be nodes, and current hierarchies are based on floors and rooms~\cite{hovsg} or a single room node~\cite{graphpad}. Using walls, doors and windows as the persistent first layer allows reporting a recovered object relative to the surface people would name; Sec.~\ref{sec:ablation} gauges the value of this option. We deliberately use off-the-shelf perception modules since the contribution is the representation, its index, and the proof that this combination can be reached from casual RGB video at $14$\,KB per scene.

In short, prior work treats geometry~\cite{colmap,mur2017orb,vggt,liu2025slam3r}, structural detection~\cite{zou2019manhattan,sceneParsing1,sceneParsing2}, open-vocabulary objects~\cite{clip,glip,dino,sam}, and episodic retrieval~\cite{threedmem} independently, but rarely integrates them into one metric hierarchy~\cite{hughes2022hydra,conceptgraphs,hovsg}. SpatialMem connects layout, objects and two-layer text in the same upright metric 3D frame within a rooted tree, providing readable, stable and low-latency relational querying from casual RGB input.

\vspace{-0.1cm}
\section{Method}
\vspace{-0.1cm}
\subsection{Problem Setup and Design Principles}
We are given a casually recorded RGB sequence $\{I_t\}$ with estimated intrinsics $K_t$ and estimated extrinsics $(R_t,t_t)$, with no dataset-provided calibration, pose or depth used during construction, and we rebuild a point set $P$ in an upright, metric coordinate frame. We aim at building a \emph{unified, metric-anchored scene memory} that can answer (i) object/location queries, (ii) relational questions (e.g., ``left of the window''), and (iii) offline guidance/retrieval over time. The memory is represented as a rooted tree $T=(V,E)$ whose typed nodes/edges are specified below. The heavy geometry is calculated once offline and the downstream retrieval and guidance functions are performed on the constructed memory at low latency. Fig.~\ref{fig:pipeline} demonstrates output on a real capture in which the pipeline reconstructs the room and its anchors in an upright metric frame, describes every object from its best viewpoints, and returns each object to its size and location ($77$ objects, $16$ walls, $2$ doors, $1$ window, $2.4$\,m ceiling). The figure also demonstrates both uses of the memory in Sec.~\ref{sec:results}, selecting the evidence frames that respond to an open-vocabulary question, where the same model fails without that evidence, and supporting metric reasoning like room area, counts and egocentric direction.
\subsection{3D Environment Preparation}
\label{sec:prep}
The minimal state that the memory requires is a view-consistent point cloud and trajectory in an upright metric frame plus stable structural seeds for Level-1. We predict the poses and depth from egocentric RGB with a swappable back-end (VGGT/SLAM3R/COLMAP; VGGT in all quantitative results), merge into a point set $P$, and align it by fitting the floor plane (normal $\rightarrow +z$), stabilizing the heading, and imposing a height prior for scale, which gives an allocentric metric frame with well-defined predicates.

We propose anchors on the aligned geometry, treating walls as large vertical planes and doors/windows as thin vertical openings, and subsequently check them across views, remove duplicates and retain only those that have enough point support, multi-view coverage and temporal persistence. The same layout pass also coarsely seeds 3D object boxes; they are provisional and will be refined further by multi-view evidence in the memory. Each of the reported configurations requires the layout model rather than treating it as optional, because the anchors it generates are the ones the hierarchy is built on. Stable anchors are promoted to Level-1 nodes with geometry \(G(v)\), semantics \(S(v)\), and confidence, ready to have objects attached; they initiate the tree and anchor the layers of \S\ref{sec:rep}.

\subsection{Unified Hierarchical Spatial Memory}
\label{sec:rep}
The hierarchy is represented by $T=(V,E)$, which is rooted and metric with four levels: The root stores the metadata of the scene and the overall frame (rooms, in case of an indoor scene). Level-1 contains the anchors, the structural components (walls, doors, windows) with oriented planes/boxes identified by SpatialLM~\cite{spatiallm} that start the tree-building process. Level-2 contains object instances connected to 3D boxes and multi-view 2D crops/masks, which are cleaned up by semantic models. Level-3 contains descriptions, including attributes (category, color, size, material) and relationships with anchors and objects, with a stable layer for consistency. All nodes $v$ contain geometry $G(v)$, semantics $S(v)$, and text $D(v)$; edges are parent--child and typed relations (near/attached/left-of).

\begin{figure}[t]
    \centering
    \includegraphics[width=\linewidth]{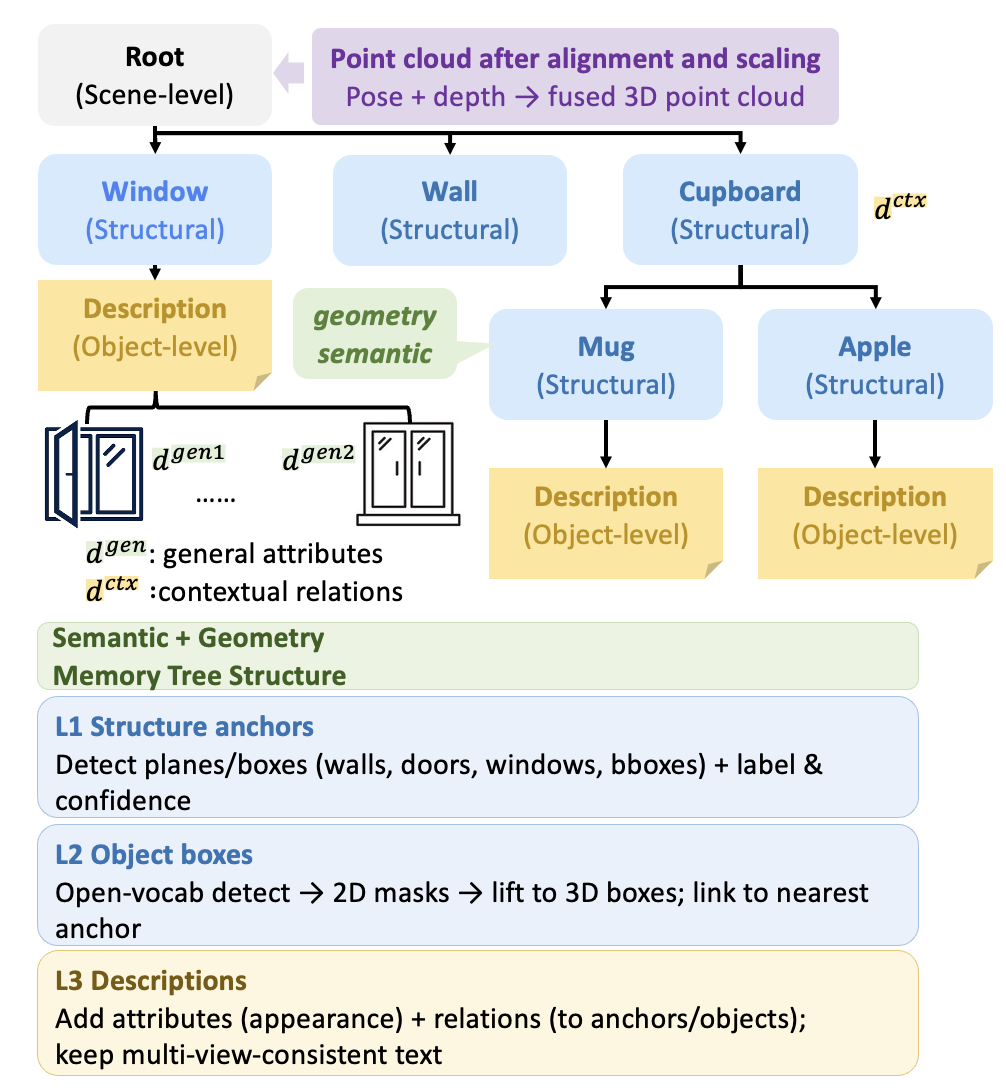}
    \caption{Tree-structured scene memory with anchors, objects, and two-layer text.}
    \label{fig:tree}
\end{figure}
\subsection{Metric Grounding and Relational Semantics}
After alignment, the frame becomes upright with $+z$ up, meaning that vertical relations are uniform. An object counts as \emph{on} a support when its bottom is level with the support's top and their footprints overlap, as \emph{above} when it is clearly higher, and as \emph{below} when clearly lower, each applied using fixed geometric heuristics with tolerance limits whose precise values are provided in Appendix~\ref{sec:supp-impl}. The lateral relations (\emph{left/right/front/behind}) are relative to the viewpoint, thus they are stored as frame tags that are resolved across views with anchors and maintained as view-local clues in case of cross-view disagreement instead of being coerced into a choice. In summary, vertical predicates are evaluated once in the $z$-aligned frame, whereas lateral predicates begin egocentrically and are merged into allocentric relations when the evidence is trustworthy.
\subsection{Two-Layer Descriptions}
The first, image-level layer is linked to the present frame. Objects are detected and segmented in 2D, lifted to 3D, and given brief descriptions that reflect the current view, so the contents of this layer may vary with occlusion, lighting or viewpoint. The scene-level layer, which comes second, summarizes what has been true across frames and promotes observations only where attributes or anchor-relative relations are consistent over repeated evidence, avoiding drift, and thus serves as the default reference point for guidance and retrieval. All relationships within the two layers are connected to named 3D anchors or objects in the common metric frame, and therefore relational queries converge to clear paths in the memory graph.

\subsection{Query and Retrieval (Low Latency)}
\label{sec:query}
Queries traverse the memory graph $T$. A ``locate'' query involves matching names and geometry, e.g.\ ``mug on the desk near the left wall'', whereas a ``relational'' query follows a short chain such as \texttt{wall} $\rightarrow$ \texttt{window} $\rightarrow$ \texttt{mug}, verifying distances and directions at every step, and guidance and retrieval convert an instruction into anchor and object waypoints on top of a prebuilt memory. Indices are based on node type and 3D region, allowing a query to touch only nearby candidates with simple distance, orientation and visibility calculations, and the indices are constructed only once, minimizing test-time latency (algorithms and measured latency in Appendices~\ref{sec:supp-algorithm} and~\ref{sec:supp-runtime}). The pipeline is model-agnostic; ingestion parameters, schedules, thresholds and per-corpus construction costs are in Appendices~\ref{sec:supp-impl} and~\ref{sec:supp-corpora}.

\section{Experimental Setup}
\subsection{Datasets}
Quantitative assessment is based on public data at benchmark scale, with each memory reconstructed using RGB images alone; a small set of casually captured egocentric scenes serves as a qualitative control (Appendix~\ref{sec:supp-egocentric}).

\begin{itemize}
  \setlength{\itemsep}{1pt}
  \item Our main evaluation is \textbf{VSI-Bench}~\cite{vsibench}, which consists of $288$ room-tour videos of ScanNet, ScanNet++ and ARKitScenes with $5{,}130$ questions over eight tasks, and we construct one memory per video from RGB only. $287$ out of $288$ scenes are reconstructed, and the failure is documented as a miss rather than a drop.
  \item \textbf{OpenEQA}~\cite{openeqa}, in its episodic-memory split, asks open-vocabulary questions which are answered based on past experience, the setting our memory is designed to serve. The ScanNet component provides $953$ questions across $79$ scenes (the $10$ missing scenes are held out in the test split); we repeat the frame baselines with the same set of questions, and published rows are just context.
  \item The \textbf{ScanNet suite} probes \emph{how} an answer is grounded, on the $12$ scenes that have both VSI-Bench coverage and complete official annotations. The ground truth is based on the labels in \texttt{aggregation.json} and the released instance boxes, and not on our reconstruction, and the scene ids are given in Appendix~\ref{sec:supp-protocol}.
  \item Two self-recorded \textbf{egocentric captures} and one Replica room serve as a qualitative probe of motion blur and clutter; their system-proposed, human-verified ground truth does not support any head-to-head statement.
\end{itemize}

The suite queries are generated deterministically from the annotations as balanced relation probes and single-instance retrieval queries; a human audit confirms $98.6\%$ of sampled episodes, rejected episodes are removed for every system alike, and the result is the $573$ relation and $471$ retrieval episodes that all tables report (protocol in Appendix~\ref{sec:supp-gt}).

Construction is a one-time cost with a median of $118$\,s per scene, and queries access only the index of layout, objects and descriptions, with a median of $14.3$\,KB per scene (about $60$ nodes; $4.07$\,MB across all $287$ scenes). Linked evidence frames, point clouds and intermediates are excluded, as they are not read by the index scan; byte-level accounting is provided in Appendix~\ref{sec:supp-footprint}.

The geometry back-end of all quantitative results is VGGT, on an \emph{offline} memory that is built only once per recorded sequence. Following pose/depth recovery we align to an upright frame and set the metric scale by a height prior. The prior is a constant $2.5$\,m ceiling height for every scene; there is no per-scene ceiling height, depth, pose or intrinsic from any dataset in the construction path. The thresholds and runtimes are provided in Appendices~\ref{sec:supp-impl} and~\ref{sec:supp-runtime}.

\subsection{Metrics}\label{sec:metrics}
Our assessments are based on three criteria; more specific definitions, thresholds, and procedures can be found in Appendix~\ref{sec:supp-metrics}. On VSI-Bench we use the metrics that the benchmark provides and verify the harness by replicating a published row. The accuracies carry $95\%$ bootstrap confidence intervals; whenever a table has point estimates, the intervals are in Appendix~\ref{sec:supp-ci}, and tables are produced directly from the stored prediction files.

The first task tests relative position, where each question lists two targets and the system responds ``$A$ is \emph{rel} of $B$'', scored against the human-verified annotation ($\mathrm{Acc}_{\text{rel}}$); on the suite these probes are balanced yes/no (chance $0.5$). Navigation-style guidance is the next task, a qualitative demonstration only, whose four metrics are kept intact in Appendix~\ref{sec:supp-nav-retained}. The third task is object retrieval, where \(\mathrm{SR}_{\text{obj}}\) determines if the target is verified within the distance criterion and \(Acc_{\text{parent}}\)/\(Acc_{\text{path}}\) whether the expected parent or full path corresponds to the annotation. Against the external baseline, parents are matched by type within $1.0$\,m for both systems alike, and the RGB-only memory incurs a scoring-time similarity penalty charged against us (details in the Appendix). All of the metrics are listed in the corpus table in Appendix~\ref{sec:supp-corpora} (Tab.~\ref{tab:corpora}), including what they measure and which evaluation reports them.

\section{Results}
\label{result}
\label{sec:results}
We report layout understanding on VSI-Bench (Tab.~\ref{tab:vsi}), open-vocabulary QA on OpenEQA (Tab.~\ref{tab:openeqa}), object retrieval and hierarchy on the $12$-scene ScanNet suite (Tab.~\ref{tab:suite}), and a per-layer ablation (Tab.~\ref{tab:ablation_main}); navigation-style guidance is shown qualitatively in Appendix~\ref{sec:supp-nav-retained}. Our harness reproduces a published row (LLaVA-NeXT-Video-7B: $35.7$ vs.\ $35.6$ overall, within $0.5$ on every task), and because our answerer does not respond to absent entities whereas the baselines always do, the suite is also rated on the matched $390$-episode set.

\subsection{Layout Understanding}
\label{sec:vsi}
\begin{table*}[!t]
\centering
\caption{\textbf{VSI-Bench} ($288$ videos, $5{,}130$ questions), scored with the benchmark's own metric code.}
\label{tab:vsi}
\scriptsize
\setlength{\tabcolsep}{3pt}
\resizebox{\textwidth}{!}{
\begin{tabular}{lccccccccccc}
\toprule
& & & \multicolumn{4}{c}{\textbf{Metric measurement}} & \multicolumn{4}{c}{\textbf{Configurational}} & \\
\cmidrule(lr){4-7}\cmidrule(lr){8-11}
\textbf{System} & \makecell{3D\\mem} & \makecell{Train-\\free} & \textbf{Cnt}$\uparrow$ & \textbf{Abs.\ Dist}$\uparrow$ & \textbf{Obj Sz}$\uparrow$ &
\textbf{Room Sz}$\uparrow$ & \textbf{Rel Dist}$\uparrow$ & \textbf{Rel Dir}$\uparrow$ & \textbf{Route}$\uparrow$ &
\textbf{Order}$\uparrow$ & \textbf{Overall}$\uparrow$ \\
\midrule
\multicolumn{12}{l}{\emph{Zero-shot, published for context}~\cite{vsibench}} \\
Frequency prior & \xmark & \cmark & 62.1 & 32.0 & 29.9 & 33.1 & 25.1 & 47.9 & 28.4 & 25.2 & 35.5 \\
GPT-4o & \xmark & \cmark & 46.2 & 5.3 & 43.8 & 38.2 & 37.0 & 41.3 & 31.5 & 28.5 & 34.0 \\
Gemini-1.5 Pro & \xmark & \cmark & 56.2 & 30.9 & 64.1 & 43.6 & 51.3 & 46.3 & 36.0 & 34.6 & 45.4 \\
LLaVA-NeXT-Video-72B & \xmark & \cmark & 48.9 & 22.8 & 57.4 & 35.3 & 42.4 & 36.7 & 35.0 & 48.6 & 40.9 \\
LLaVA-NeXT-Video-7B & \xmark & \cmark & 48.5 & 14.0 & 47.8 & 24.2 & 43.5 & 42.4 & 34.0 & 30.6 & 35.6 \\
\midrule
\multicolumn{12}{l}{\emph{Spatially instruction-tuned, published (overall only): Spatial-MLLM-4B 47.0, VG-LLM-4B 47.3, VLM-3R~\cite{vlm3r} 60.9}} \\
\midrule
\multicolumn{12}{l}{\emph{This work}, same harness, training-free} \\
LLaVA-NeXT-Video-7B, our re-run & \xmark & \cmark & \textbf{\underline{48.6}} & \underline{13.9} & 47.6 & \underline{24.4} & \textbf{43.9} & \underline{42.6} & \textbf{33.5} & 31.1 & 35.7 \\
SpatialMem w/o object nodes & \cmark & \cmark & \underline{13.9} & \textbf{\underline{28.9}} & \textbf{54.5} & \textbf{41.4} & 32.5 & \textbf{\underline{44.5}} & 31.4 & 28.5 & \underline{34.5} \\
\textbf{SpatialMem (Ours)} & \cmark & \cmark & \underline{24.9} & \underline{27.5} & 52.1 & \textbf{41.4} & 36.1 & \textbf{\underline{44.5}} & 32.5 & \textbf{41.1} & \textbf{37.5} \\
\cmidrule(lr){1-12}
\emph{$\Delta$ object layer} & & & \emph{+11.0} & \emph{$-1.4$} & \emph{$-2.4$} & \emph{$\pm0.0$} & \emph{+3.6} & \emph{$\pm0.0$} & \emph{+1.1} & \emph{+12.6} & \emph{+3.0} \\
\bottomrule
\end{tabular}}
% \vspace{-0.4cm}
\end{table*}

SpatialMem scores $37.5$ overall ($95\%$ CI $[36.1,38.8]$) on $288$ public room-tour videos (Tab.~\ref{tab:vsi}), which is higher than the frequency prior and GPT-4o, statistically similar to the $7$B video model and lower than the $72$B and proprietary models. In the table, \emph{3D mem} indicates a persistent 3D memory that remains after the query and \emph{Train-free} indicates no task-specific spatial training; \textbf{bold} is the highest among our rows and \underline{underline} is lower than the prior on that task. The instruction-tuned block is divided by training data instead of representation, because VLM-3R uses the same $7$B base as our re-run row, from $35.6$ to $60.9$, with $200$K spatial QA pairs. The per-task profile is what matters. The memory wins where a stored inventory should, on the size of the rooms and the order of appearance, and loses where the coverage limits it, as it does most obviously in counting, where an incomplete inventory fares worse than guessing the modal count. They are both constrained by one quantity, namely the observation that only $56.3\%$ of entities queried can be resolved into a memory object, and not by the metric reasoning built on top.

The contribution of the object layer can be read directly from the table. The middle row of the \emph{This work} block deletes all object nodes with a single switch, and the last row is the total difference of $3.0$ points where the intervals do not overlap, taking the system below the prior; the two tasks that \emph{benefit} are those where a detection box is coarser than the layout's extent, and therefore we exclude detection boxes from extent questions with fewer than three supporting views. The resolved-entity split (Tab.~\ref{tab:resolved}) indicates which gains are real; in each case room size is calculated using stored geometry, but absolute distance resolves for only $30\%$ of the questions and performs no better on them, so we discard it as evidence of metric grounding.

\subsection{Open-Vocabulary QA from Memory}
\label{sec:openeqa}
\begin{table*}[!t]
\centering
\caption{\textbf{OpenEQA, episodic-memory task}, LLM-Match with a GPT-4o judge.}
\label{tab:openeqa}
\scriptsize
\setlength{\tabcolsep}{3pt}
\resizebox{\textwidth}{!}{
\begin{tabular}{llccccccccccc}
\toprule
\textbf{System} & \textbf{Memory use} & \makecell{Poses\\given} & \makecell{Frames\\@query} & \textbf{Text}
& ObjRec & Attr & State & Loc & Spatial & Func & World & \textbf{Overall}$\uparrow$ \\
\midrule
\multicolumn{13}{l}{\emph{Published for context, full EM-EQA set ($1{,}636$ q): blind GPT-4$^{\dagger}$ 35.5, CG captions$^{\dagger}$~\cite{conceptgraphs} 34.4, GPT-4V@50$^{\dagger}$ 48.1, human 85.9~\cite{openeqa}}} \\
VL-KnG~\cite{vlkng} & graph retrieval & \xmark & --- & \cmark & --- & --- & --- & --- & --- & --- & --- & 55.2 \\
GraphPad~\cite{graphpad} & keyframe retrieval & \req & $5$ & \cmark & --- & --- & --- & --- & --- & --- & --- & 55.3 \\
3D-Mem~\cite{threedmem} & snapshot retrieval & \req & $3.1$ & \xmark & --- & --- & --- & --- & --- & --- & --- & 57.2 \\
\midrule
\multicolumn{13}{l}{\emph{This work: full ScanNet portion, $953$ questions / $79$ scenes, identical questions, answerer and judge}} \\
Blind GPT-4o (control) & none & \xmark & $0$ & \xmark & 32.9 & 40.2 & 57.9 & 34.8 & 41.2 & 58.7 & 47.2 & 44.5 \\
SpatialMem, serialised text & serialised & \xmark & $0$ & \cmark & 21.9 & 30.3 & 55.3 & 21.3 & 33.4 & 47.1 & 37.6 & \underline{35.1} \\
\quad w/o object nodes & serialised & \xmark & $0$ & \cmark & 12.2 & 26.7 & 50.9 & 21.1 & 29.7 & 31.8 & 30.2 & \underline{29.0} \\
\quad w/o descriptions$^{\S}$ & serialised & \xmark & $0$ & \cmark & 16.0 & 33.2 & 48.4 & 20.8 & 32.8 & 40.5 & 33.9 & \underline{32.0} \\
GPT-4o, uniform frames & none & \xmark & $8$ & \xmark & 60.8 & 70.4 & 73.9 & 46.3 & 54.9 & 67.8 & 56.6 & 61.6 \\
GPT-4o, uniform frames & none & \xmark & $50$ & \xmark & 73.2 & \textbf{80.7} & 79.4 & \textbf{56.7} & \textbf{63.1} & \textbf{73.4} & 68.9 & 70.9 \\
Gemini-2.5-Flash, uniform frames & none & \xmark & $50$ & \xmark & \textbf{74.8} & 78.9 & \textbf{86.3} & 55.9 & 61.5 & 73.2 & \textbf{73.8} & \textbf{72.1} \\
Qwen2.5-VL-72B, uniform frames & none & \xmark & $50$ & \xmark & 63.4 & 74.5 & 78.3 & 47.0 & 50.4 & 66.0 & 61.0 & 63.1 \\
\textbf{SpatialMem retrieval (Ours)} & \textbf{evidence-frame retrieval} & \xmark & $8$ mem. & \xmark & 62.6 & 68.3 & 74.3 & 48.0 & 55.4 & 67.0 & 57.8 & 62.0 \\
\quad + serialised text$^{\S}$ & retrieval + serialised & \xmark & $8$ mem. & \cmark & 53.7 & 74.5 & 79.3 & 43.8 & 41.1 & 58.9 & 50.0 & 57.6 \\
\bottomrule
\end{tabular}}
% \vspace{-0.4cm}
\end{table*}

In order to test the same memory with open-vocabulary questions written by humans, we assess the episodic-memory task of OpenEQA~\cite{openeqa}, which requires a system to respond based on a recollection of past experience, the basis of our memory, and the one public place where \emph{other persistent memories} report numbers. Tab.~\ref{tab:openeqa} puts our representation next to a caption-serialised scene graph~\cite{conceptgraphs}, an editable scene graph~\cite{graphpad}, a knowledge graph without 3D reconstruction~\cite{vlkng}, and a snapshot memory~\cite{threedmem}; two out of four need posed RGB-D, whereas ours does not.

The ScanNet part of the episodic-memory split ($953$ questions, $79$ scenes) is measured using the benchmark's LLM-Match judge and scale. In Tab.~\ref{tab:openeqa} the top block is published on the whole set and is context only ($^{\dagger}$as shown in~\cite{threedmem}), and the bottom block is our harness with the same questions, answerer (GPT-4o) and judge for each row, and the one we compare within ($^{\S}$on the $320$-question subset constructed first). \emph{Text} marks inputs that contain the serialised memory, and \emph{mem.}\ marks frames determined by its evidence links. The blind row, which is exposed to the question and no visual input, is used as a control on language priors.

However, what is important is not whether the memory is present but rather the manner in which it is consumed. When serialised to text, the memory scores $35.1$, \emph{nine points lower than the blind control} of $44.5$, matching the published pattern where the caption-serialised representation is lower than blind (CG captions $34.4$ vs.\ $35.5$); fixing a well-developed model to a small inventory eliminates the priors that address the appearance-heavy categories of OpenEQA. All the systems published so far that beat blind give the answerer \emph{pixels}, and the memory's role there is to select them. Applied in that sense, matching the entities of the question to stored evidence frames with $K{=}8$ under the benchmark's own frame prompt, the very same memory scores $\mathbf{62.0}$ $[59.7,64.3]$. The controlled comparison is uniform sampling with the same answerer and frame budget, which is equal to $61.6$; in these single-room captures eight uniform frames are enough to cover the room, so retrieval coincides with uniform selection instead of beating it, and the memory regains frame-level accuracy while supporting every answer with named 3D nodes and associated evidence. The retrieval row sits above blind precisely on the memory categories (localisation $48.0$ vs.\ $34.8$, spatial $55.4$ vs.\ $41.2$), matches the performance of $50$-frame Qwen2.5-VL-72B at a sixth of the frames, and is $9$--$10$ below $50$-frame answering. A sweep of the budget clarifies this reading. Keeping the answerer constant and reducing $K$, memory selection remains better than uniform sampling with a gap that increases as the budget tightens, from $+0.4$ at $K{=}8$ to $+1.0$ at $K{=}4$ and $+2.6$ at $K{=}2$ ($52.7$ vs.\ $50.0$, surpassing uniform's upper confidence bound), and it is particularly high on localisation ($43.8$ vs.\ $37.4$ at $K{=}2$). The choice is relevant specifically in cases of few frames. Adding the serialised text \emph{alongside} the same frames results in a cost of $4.9$ points, which completes the $2{\times}2$ separating retrieval (neutral-to-positive) from serialisation (harmful).

\begin{table*}[!t]
\caption{\textbf{12-scene ScanNet suite}, all tasks scored against \emph{official} instance annotations.}
\label{tab:suite}
\centering
\scriptsize
\setlength{\tabcolsep}{3pt}
\resizebox{\textwidth}{!}{
\begin{tabular}{llccccccccccc}
\toprule
& & & \multicolumn{2}{c}{\textbf{Relation verif.}} & \multicolumn{4}{c}{\textbf{Object retrieval}} & \multicolumn{2}{c}{\textbf{Hierarchy}} & \textbf{Descr.} & \multicolumn{1}{c}{\textbf{Cost}} \\
\cmidrule(lr){4-5}\cmidrule(lr){6-9}\cmidrule(lr){10-11}\cmidrule(lr){12-12}\cmidrule(lr){13-13}
\textbf{Model} & \textbf{Input} & \makecell{Poses\\given}
& all$\uparrow$ & matched$\uparrow$
& SR@$0.5\uparrow$ & SR@$1.0\uparrow$ & mLE\,[m]$\downarrow$ & R@$3\uparrow$
& Wall$\uparrow$ & All$\uparrow$ & Colour$\uparrow$ & \makecell{Build\\s/scene$\downarrow$} \\
& & & \tiny(573) & \tiny(390) & \tiny(471) & \tiny(471) & \tiny(471) & \tiny(471) & \tiny(436) & \tiny(471) & \tiny(19) & \\
\midrule
GPT-4o           & 8 frames & \xmark & \textbf{0.61} & 0.615 & \multicolumn{6}{c}{\multirow{4}{*}{\emph{returns no 3D instance}}} & 0.74 & --- \\
Gemini-2.5-Flash & 8 frames & \xmark & 0.60 & \textbf{0.623} & \multicolumn{6}{c}{} & 0.79 & --- \\
Qwen2.5-VL-72B   & 8 frames & \xmark & 0.59 & 0.610 & \multicolumn{6}{c}{} & 0.74 & --- \\
GPT-4o-mini      & 8 frames & \xmark & 0.55 & 0.549 & \multicolumn{6}{c}{} & \textbf{0.84} & --- \\
\midrule
ConceptGraphs~\cite{conceptgraphs} & posed RGB-D & \req & \textbf{0.61} & \textbf{0.623} & \textbf{0.41} & \textbf{0.50} & \textbf{1.03} & --- & 0.00 & 0.01 & n/a & 234 \\
\midrule
\textbf{SpatialMem-RGBD} & posed RGB-D & \req & 0.55 & \textbf{0.667} & 0.26 & 0.36 & 1.74 & \textbf{0.40} & \textbf{0.25} & \textbf{0.24} & n/a & \textbf{89} \\
\textbf{SpatialMem (Ours)} & RGB video & \xmark & 0.43 & 0.610 & 0.18 & 0.24 & 2.14 & 0.22 & 0.16 & 0.15 & 0.42 & 118 \\
\bottomrule
\end{tabular}}
\end{table*}

\begin{table*}[!t]
\centering
\caption{\textbf{Contribution of each memory layer}, reported as the change each removal causes.}
\label{tab:ablation_main}
\scriptsize
\setlength{\tabcolsep}{3pt}
\resizebox{\textwidth}{!}{
\begin{tabular}{llccccccccc}
\toprule
& & \multicolumn{6}{c}{\textbf{ScanNet suite} \tiny($573$ relation / $471$ retrieval, external GT)} & \multicolumn{3}{c}{\textbf{VSI-Bench} \tiny($5{,}130$)} \\
\cmidrule(lr){3-8}\cmidrule(lr){9-11}
\textbf{Component removed or replaced} & \textbf{Module}
& Coverage & Ver.\ all & Ver.\ answ. & SR@$0.5$ & SR@$1.0$ & $\mathrm{Acc_{parent}}$
& \makecell{Entity\\cov.} & Metric & Config. \\
\midrule
\emph{reference}: full RGB memory & --- & \emph{0.764} & \emph{0.433} & \emph{0.611} & \emph{0.176} & \emph{0.236} & \emph{0.149} & \emph{56.3} & \emph{36.5} & \emph{38.5} \\
\midrule
Estimated geometry (vs.\ given posed depth) & geometry & $-0.108$ & $-0.120$ & $-0.051$ & $-0.087$ & $-0.123$ & $-0.093$ & --- & --- & --- \\
Scale $-10\%$ & geometry & $0.000$ & $-0.004$ & $-0.005$ & $0.000$ & $0.000$ & $-0.005$ & --- & --- & --- \\
Scale $+10\%$ & geometry & $0.000$ & $-0.007$ & $-0.010$ & $-0.013$ & $-0.019$ & $-0.009$ & --- & --- & --- \\
w/o open-vocabulary object nodes & object layer & $\mathbf{-0.573}$ & $\mathbf{-0.354}$ & $+0.006$ & $-0.149$ & $-0.193$ & $-0.115$ & $\mathbf{-25.5}$ & $-1.8$ & $-4.3$ \\
w/o wall anchors & anchors & $0.000$ & $0.000$ & $0.000$ & $0.000$ & $0.000$ & $\mathbf{-0.130}$ & --- & --- & --- \\
w/o structural layer (flat) & anchors & $0.000$ & $0.000$ & $0.000$ & $0.000$ & $0.000$ & $\mathbf{-0.149}$ & --- & --- & --- \\
w/o descriptions / Layer-1 only & descriptions & \multicolumn{6}{c}{\emph{not consumed by the suite's answering path}} & --- & --- & --- \\
w/o relational hint in retrieval & retrieval & $0.000$ & $0.000$ & $0.000$ & $-0.002$ & $+0.002$ & $+0.008$ & --- & --- & --- \\
\bottomrule

\end{tabular}}
% \vspace{-0.2cm}
\end{table*}

% \subsection{Indoor Guidance}
% \label{sec:guidance}
% Since this task is not autonomous navigation but rather guidance over a prebuilt reference, and since its ground truth is system-proposed and then human-verified, it has become a \emph{qualitative demonstration}, and its four metrics have been preserved in the Supplementary Material with no quantitative assertion based on them.

\subsection{Object Retrieval Task Analysis}
\label{sec:suite}

Tab.~\ref{tab:suite} reports retrieval, hierarchy and description quality at four levels of input. Frame models take eight frames and no 3D; our memory consists of RGB video only; SpatialMem-RGBD builds the same representation from the dataset's posed depth, isolating the geometry front end; and ConceptGraphs~\cite{conceptgraphs} is given ground-truth poses, intrinsics and axis alignment, hence its map is in the evaluation frame whereas our RGB row incurs a scoring-time alignment (a per-scene 4-DoF similarity fitted only at scoring time, outside the answering path). Ground truth comes from the official annotations, irrespective of any reconstruction; parenthesised counts are audited episodes, and per-predicate and Door/Window breakdowns are in the Appendix.

Relation verification does not split the systems, since all seven lie between $0.55$ and $0.67$ and their intervals overlap; a relationship between two named objects based on properly framed perspectives of a small room can be answered mostly by appearance. What an explicit memory buys can be seen in the other groups.

Retrieval and hierarchy divide the systems in opposite directions. The privileged baseline has better localization, and two additional columns indicate where the miss rate is located. mLE, the median distance between the returned instance and the annotated centre, sorts the systems by input ($1.03$/$1.74$/$2.14$\,m), with part of the RGB gap being the scoring-time alignment that the baseline is spared. R@$3$ decouples ranking and selection; with matched input a correct instance can be found in the top three in $0.40$ of episodes, as frequently as the baseline's top-$1$ is correct, but raising it to rank one provides merely $+0.002$, so the failure lies in telling nearly identical instances apart at metre scale. Hierarchy runs in the opposite direction, and the parent-type split makes the mechanism explicit. Walls do not participate in a standard scene graph, whereas $436$ out of $471$ annotated parents \emph{are} walls, and the baseline's accuracy there is precisely $0.00$. On the small door and window subsets the ordering is mixed, as doors are more favourable towards the baseline ($0.21$ vs.\ $0.12$, $n{=}24$) and windows towards us ($0.18$ vs.\ $0.00$, $n{=}11$), and thus the exact claim is that object-only graphs cannot ground objects to structural parents at all, not that our anchors are better localised. Description quality works to our disadvantage (Sec.~\ref{sec:ablation}).

\subsection{Ablation Study}
\label{sec:ablation}

Tab.~\ref{tab:ablation_main} ablates each memory layer on the annotation-derived suite and on VSI-Bench. Each variant is one switch across shared caches, rated on the same episodes, with the same matched object and scoring-time transform; the coverage column is the percentage of episodes responded to without abstaining, read against answerable-only accuracy to isolate coverage losses from reasoning losses. The descriptions are not read by the answering path of the suite, and hence their rows cannot shift these columns; their value is measured at the point of their consumption, which is in Tab.~\ref{tab:openeqa} and the colour metric of Tab.~\ref{tab:suite}. All layers shift a distinct column.

Geometry is the only component whose replacement degrades every column, which is the price of RGB video. The object-node row is the most informative; eliminating the layer removes three quarters of coverage ($0.764\to0.191$) and most of retrieval, but accuracy on still-answerable episodes remains the same, thus the layer purchases \emph{coverage} and not reasoning, the same conclusion Tab.~\ref{tab:vsi} reaches from the other direction. Structural rows shift hierarchy in isolation; limiting the parents to doors/windows yields $0.019$, which replicates the baseline's $0.01$ inside our own system, a representational decision rather than a weaker implementation. The relational cue does not move anything out of noise, and descriptions only cover $34$ out of $471$ targets, therefore we re-scope them to attribute binding. A $\pm10\%$ scale perturbation changes at most $0.02$ since the scoring-time fit absorbs the global scale; fixing that fit costs SR@$0.5$ some $0.03$.

% \vspace{-0.1cm}

\subsection{Limitations}
% \vspace{-0.1cm}
We quantify three limitations. The system is bounded by coverage and not metric reasoning, because only $56\%$ of queried entities can be resolved to a memory object, and Tab.~\ref{tab:ablation_main} indicates that the missing accuracy depends on restoring the coverage. Estimated geometry costs localisation ($0.41$ vs.\ $0.18/0.26$ against the privileged-input graph), around $40\%$ of it being scoring-time alignment. Lastly, construction is done offline, the scenes are static, guidance consists of decision-point instructions over a prebuilt reference, and counting remains our weakest task. Privacy considerations are also raised by persistent memories of indoor spaces, and the deployment safeguard is on-device construction with opt-in capture.
\section{Conclusion}
We have introduced SpatialMem, a single queryable memory that is created using egocentric RGB only and incorporates an upright metric structure, anchor-attached open-vocabulary objects, and two-layer descriptions. It scores $37.5$ on VSI-Bench with a task profile earned where a stored inventory ought to succeed, regains frame-level accuracy on OpenEQA as evidence-frame retrieval that outperforms uniform sampling as the budget diminishes while serialisation falls under the blind control, and places objects under structural parents where a standard scene graph cannot ($0.24$ compared to $0.01$), with ablation identifying object coverage as the bottleneck. We release the audited suite, the probe set, and all the memories and code. We consider SpatialMem a realistic move towards spatially grounded long-horizon video understanding, with more heterogeneous scenes, gradual updates, and richer interactions as future work.

{
    \small
    \bibliographystyle{ieeenat_fullname}
    \bibliography{main}

@String(CVPR= {IEEE Conf. Comput. Vis. Pattern Recog.})

@String(ICCV= {Int. Conf. Comput. Vis.})

@String(ECCV= {Eur. Conf. Comput. Vis.})

@String(CVPR  = {CVPR})

@String(ICCV  = {ICCV})

@String(ECCV  = {ECCV})

@article{colmap,
  title={Structure-from-Motion Revisited},
  author={Johannes L. Sch{\"o}nberger and Jan-Michael Frahm},
  journal={2016 IEEE Conference on Computer Vision and Pattern Recognition (CVPR)},
  year={2016},
  pages={4104-4113},
  url={https://api.semanticscholar.org/CorpusID:1728538}
}

@article{mur2017orb,
  title={Orb-slam2: An open-source slam system for monocular, stereo, and rgb-d cameras},
  author={Mur-Artal, Raul and Tard{\'o}s, Juan D},
  journal={IEEE transactions on robotics},
  volume={33},
  number={5},
  pages={1255--1262},
  year={2017},
  publisher={IEEE}
}

@inproceedings{nerf,
  title={NeRF: Representing scenes as neural radiance ﬁelds for view synthesis},
  author={Ben Mildenhall and Pratul P. Srinivasan and Matthew Tancik and Jonathan T. Barron and Ravi Ramamoorthi and Ren Ng},
  booktitle={European Conference on Computer Vision (ECCV)},
  year={2020}
}

@article{vggt,
  title={VGGT: Visual Geometry Grounded Transformer},
  author={Jianyuan Wang and Minghao Chen and Nikita Karaev and Andrea Vedaldi and Christian Rupprecht and David Novotn{\'y}},
  journal={2025 IEEE/CVF Conference on Computer Vision and Pattern Recognition (CVPR)},
  year={2025},
  pages={5294-5306},
  url={https://api.semanticscholar.org/CorpusID:277043968}
}

@article{liu2025slam3r,
  title={SLAM3R: Real-Time Dense Scene Reconstruction from Monocular RGB Videos},
  author={Yuzheng Liu and Siyan Dong and Shuzhe Wang and Yanchao Yang and Qingnan Fan and Baoquan Chen},
  journal={2025 IEEE/CVF Conference on Computer Vision and Pattern Recognition (CVPR)},
  year={2025},
  pages={16651-16662},
  url={https://api.semanticscholar.org/CorpusID:274655772}
}

@article{sceneParsing1,
  title={FRNet: Feature Reconstruction Network for RGB-D Indoor Scene Parsing},
  author={Wujie Zhou and Enquan Yang and Jingsheng Lei and Lu Yu},
  journal={IEEE Journal of Selected Topics in Signal Processing},
  year={2022},
  volume={16},
  pages={677-687},
  url={https://api.semanticscholar.org/CorpusID:248770222}
}

@article{sceneParsing2,
  title={PGDENet: Progressive Guided Fusion and Depth Enhancement Network for RGB-D Indoor Scene Parsing},
  author={Wujie Zhou and Enquan Yang and Jingsheng Lei and Jian Wan and Lu Yu},
  journal={IEEE Transactions on Multimedia},
  year={2023},
  volume={25},
  pages={3483-3494},
  url={https://api.semanticscholar.org/CorpusID:247637924}
}

@article{clip,
  title={Learning Transferable Visual Models From Natural Language Supervision},
  author={Alec Radford and Jong Wook Kim and Chris Hallacy and Aditya Ramesh and Gabriel Goh and Sandhini Agarwal and Girish Sastry and Amanda Askell and Pamela Mishkin and Jack Clark and Gretchen Krueger and Ilya Sutskever},
  journal={ArXiv},
  year={2021},
  volume={abs/2103.00020},
  url={https://api.semanticscholar.org/CorpusID:231591445}
}

@article{glip,
  title={Grounded Language-Image Pre-training},
  author={Liunian Harold Li and Pengchuan Zhang and Haotian Zhang and Jianwei Yang and Chunyuan Li and Yiwu Zhong and Lijuan Wang and Lu Yuan and Lei Zhang and Jenq-Neng Hwang and Kai-Wei Chang and Jianfeng Gao},
  journal={2022 IEEE/CVF Conference on Computer Vision and Pattern Recognition (CVPR)},
  year={2022},
  pages={10955-10965},
  url={https://api.semanticscholar.org/CorpusID:244920947}
}

@article{dino,
  title={Grounding DINO: Marrying DINO with Grounded Pre-Training for Open-Set Object Detection},
  author={Shilong Liu and Zhaoyang Zeng and Tianhe Ren and Feng Li and Hao Zhang and Jie Yang and Chun-yue Li and Jianwei Yang and Hang Su and Jun-Juan Zhu and Lei Zhang},
  journal={ArXiv},
  year={2023},
  volume={abs/2303.05499},
  url={https://api.semanticscholar.org/CorpusID:257427307}
}

@article{sam,
  title={Segment Anything},
  author={Alexander Kirillov and Eric Mintun and Nikhila Ravi and Hanzi Mao and Chlo{\'e} Rolland and Laura Gustafson and Tete Xiao and Spencer Whitehead and Alexander C. Berg and Wan-Yen Lo and Piotr Doll{\'a}r and Ross B. Girshick},
  journal={2023 IEEE/CVF International Conference on Computer Vision (ICCV)},
  year={2023},
  pages={3992-4003},
  url={https://api.semanticscholar.org/CorpusID:257952310}
}

@article{imgCaption,
  title={Deep Learning Approaches on Image Captioning: A Review},
  author={Taraneh Ghandi and Hamid Reza Pourreza and Hamidreza Mahyar},
  journal={ACM Computing Surveys},
  year={2022},
  volume={56},
  pages={1 - 39},
  url={https://api.semanticscholar.org/CorpusID:246430542}
}

@misc{LLMparsing,
      title={LLM-Guided Indoor Navigation with Multimodal Map Understanding}, 
      author={Alberto Coffrini and Paolo Barsocchi and Francesco Furfari and Antonino Crivello and Alessio Ferrari},
      year={2025},
      eprint={2503.11702},
      archivePrefix={arXiv},
      primaryClass={cs.AI},
      url={https://arxiv.org/abs/2503.11702}, 
}

@article{zou2019manhattan,
  title={Manhattan Room Layout Reconstruction from a Single $360^{\circ}$ Image: A Comparative Study of State-of-the-Art Methods},
  author={Chuhang Zou and Jheng-Wei Su and Chi-Han Peng and Alex Colburn and Qi Shan and Peter Wonka and Hung-Kuo Chu and Derek Hoiem},
  journal={Int. J. Comput. Vis.},
  year={2021},
  volume={129},
  pages={1410-1431},
  url={https://api.semanticscholar.org/CorpusID:233978120}
}

@article{hughes2022hydra,
  title={Hydra: A Real-time Spatial Perception System for 3D Scene Graph Construction and Optimization},
  author={Nathan Hughes and Yun Chang and Luca Carlone},
  journal={Robotics: Science and Systems XVIII},
  year={2022},
  url={https://api.semanticscholar.org/CorpusID:248913107}
}

@article{blukis2022persistent,
  title={A Persistent Spatial Semantic Representation for High-level Natural Language Instruction Execution},
  author={Valts Blukis and Chris Paxton and Dieter Fox and Animesh Garg and Yoav Artzi},
  journal={ArXiv},
  year={2021},
  volume={abs/2107.05612},
  url={https://api.semanticscholar.org/CorpusID:235795722}
}

@article{macario2022comprehensive,
  title={A Comprehensive Survey of Visual SLAM Algorithms},
  author={Andr{\'e}a Macario Barros and Maugan Michel and Yoann Moline and Gwenol{\'e} Corre and Fr{\'e}d{\'e}rick Carrel},
  journal={Robotics},
  year={2022},
  volume={11},
  pages={24},
  url={https://api.semanticscholar.org/CorpusID:246794074}
}

@article{zou2022mononeuralfusion,
  title={MonoNeuralFusion: Online Monocular Neural 3D Reconstruction with Geometric Priors},
  author={Zi-Xin Zou and Shi-Sheng Huang and Yan-Pei Cao and Tai-Jiang Mu and Ying Shan and Hongbo Fu},
  journal={ArXiv},
  year={2022},
  volume={abs/2209.15153},
  url={https://api.semanticscholar.org/CorpusID:252668762}
}

@misc{zhu2024survey,
      title={A Survey on Open-Vocabulary Detection and Segmentation: Past, Present, and Future}, 
      author={Chaoyang Zhu and Long Chen},
      year={2024},
      eprint={2307.09220},
      archivePrefix={arXiv},
      primaryClass={cs.CV},
      url={https://arxiv.org/abs/2307.09220}, 
}

@misc{zhang2025survey,
      title={A Survey on the Memory Mechanism of Large Language Model based Agents}, 
      author={Zeyu Zhang and Xiaohe Bo and Chen Ma and Rui Li and Xu Chen and Quanyu Dai and Jieming Zhu and Zhenhua Dong and Ji-Rong Wen},
      year={2024},
      eprint={2404.13501},
      archivePrefix={arXiv},
      primaryClass={cs.AI},
      url={https://arxiv.org/abs/2404.13501}, 
}

@misc{conceptgraphs,
      title={ConceptGraphs: Open-Vocabulary 3D Scene Graphs for Perception and Planning}, 
      author={Qiao Gu and Alihusein Kuwajerwala and Sacha Morin and Krishna Murthy Jatavallabhula and Bipasha Sen and Aditya Agarwal and Corban Rivera and William Paul and Kirsty Ellis and Rama Chellappa and Chuang Gan and Celso Miguel de Melo and Joshua B. Tenenbaum and Antonio Torralba and Florian Shkurti and Liam Paull},
      year={2023},
      eprint={2309.16650},
      archivePrefix={arXiv},
      primaryClass={cs.RO},
      url={https://arxiv.org/abs/2309.16650}, 
}

@misc{vsibench,
      title={Thinking in Space: How Multimodal Large Language Models See, Remember, and Recall Spaces}, 
      author={Jihan Yang and Shusheng Yang and Anjali W. Gupta and Rilyn Han and Li Fei-Fei and Saining Xie},
      year={2025},
      eprint={2412.14171},
      archivePrefix={arXiv},
      primaryClass={cs.CV},
      url={https://arxiv.org/abs/2412.14171}, 
}

@inproceedings{hovsg,
    series={RSS2024},
   title={Hierarchical Open-Vocabulary 3D Scene Graphs for Language-Grounded Robot Navigation},
   url={http://dx.doi.org/10.15607/RSS.2024.XX.077},
   DOI={10.15607/rss.2024.xx.077},
   booktitle={Robotics: Science and Systems XX},
   publisher={Robotics: Science and Systems Foundation},
   author={Werby, Abdelrhman and Huang, Chenguang and Büchner, Martin and Valada, Abhinav and Burgard, Wolfram},
   year={2024},
   month=jul, collection={RSS2024} }

@misc{threedmem,
      title={3D-Mem: 3D Scene Memory for Embodied Exploration and Reasoning}, 
      author={Yuncong Yang and Han Yang and Jiachen Zhou and Peihao Chen and Hongxin Zhang and Yilun Du and Chuang Gan},
      year={2025},
      eprint={2411.17735},
      archivePrefix={arXiv},
      primaryClass={cs.CV},
      url={https://arxiv.org/abs/2411.17735}, 
}

@inproceedings{openeqa,
  title     = {OpenEQA: Embodied Question Answering in the Era of Foundation Models},
  author    = {Arjun Majumdar and Anurag Ajay and Xiaohan Zhang and Pranav Putta and Sriram Yenamandra and Mikael Henaff and Sneha Silwal and Paul Mcvay and Oleksandr Maksymets and Sergio Arnaud and Karmesh Yadav and Qiyang Li and Ben Newman and Mohit Sharma and Vincent Berges and Shiqi Zhang and Pulkit Agrawal and Yonatan Bisk and Dhruv Batra and Mrinal Kalakrishnan and Franziska Meier and Chris Paxton and Alexander Sax and Aravind Rajeswaran},
  booktitle = {2024 IEEE/CVF Conference on Computer Vision and Pattern Recognition (CVPR)},
  year      = {2024},
  doi       = {10.1109/cvpr52733.2024.01560},
}

@article{graphpad,
  title={GraphPad: Inference-Time 3D Scene Graph Updates for Embodied Question Answering},
  author={Muhammad Qasim Ali and Saeejith Nair and Alexander Wong and Yuchen Cui and Yuhao Chen},
  journal={ArXiv},
  year={2025},
  volume={abs/2506.01174},
  url={https://api.semanticscholar.org/CorpusID:279075267}
}

@inproceedings{vlkng,
  title={Spatiotemporal Knowledge Graphs as Persistent Scene Memory for Embodied Question Answering},
  author={Mohamad Al Mdfaa and Svetlana Lukina and Timur Akhtyamov and Arthur Nigmatzyanov and Dmitrii Nalberskii and Sergey Zagoruyko and Gonzalo Ferrer},
  year={2025},
  url={https://api.semanticscholar.org/CorpusID:281724788}
}

@article{openmask3d,
  title={OpenMask3D: Open-Vocabulary 3D Instance Segmentation},
  author={Ayca Takmaz and Elisabetta Fedele and Robert W. Sumner and Marc Pollefeys and Federico Tombari and Francis Engelmann},
  journal={ArXiv},
  year={2023},
  volume={abs/2306.13631},
  url={https://api.semanticscholar.org/CorpusID:259243888}
}

@article{openscene,
  title={OpenScene: 3D Scene Understanding with Open Vocabularies},
  author={Songyou Peng and Kyle Genova and ChiyuMaxJiang and Andrea Tagliasacchi and Marc Pollefeys and Thomas A. Funkhouser},
  journal={2023 IEEE/CVF Conference on Computer Vision and Pattern Recognition (CVPR)},
  year={2023},
  pages={815-824},
  url={https://api.semanticscholar.org/CorpusID:254044069}
}

@article{vlm3r,
  title={VLM-3R: Vision-Language Models Augmented with Instruction-Aligned 3D Reconstruction},
  author={Zhiwen Fan and Jian Zhang and Renjie Li and Junge Zhang and Runjin Chen and Hezhen Hu and Kevin Wang and Huaizhi Qu and Dilin Wang and Zhicheng Yan and Hongyu Xu and Justin Theiss and Tianlong Chen and Jiachen Li and Zhengzhong Tu and Zhangyang Wang and Rakesh Ranjan},
  journal={ArXiv},
  year={2025},
  volume={abs/2505.20279},
  url={https://api.semanticscholar.org/CorpusID:279449925}
}

@article{spatiallm,
  title={SpatialLM: Training Large Language Models for Structured Indoor Modeling},
  author={Yongsen Mao and Junhao Zhong and Chuan Fang and Jia Zheng and Rui Tang and Hao Zhu and Ping Tan and Zihan Zhou},
  journal={ArXiv},
  year={2025},
  volume={abs/2506.07491},
  url={https://api.semanticscholar.org/CorpusID:279251798}
}
}

% Supplementary material, included as appendices so it ships in the same arXiv PDF
\clearpage
\appendix
\section{Pipeline Visualizations}
\label{sec:supp-pipeline}

Figure~\ref{fig:supp_pipeline} gives a comprehensive view of the entire SpatialMem pipeline at high resolution. It shows the steps of the process starting with egocentric RGB input up to a searchable 3D memory tree:

\begin{figure*}[t]
    \centering
    \includegraphics[width=\linewidth]{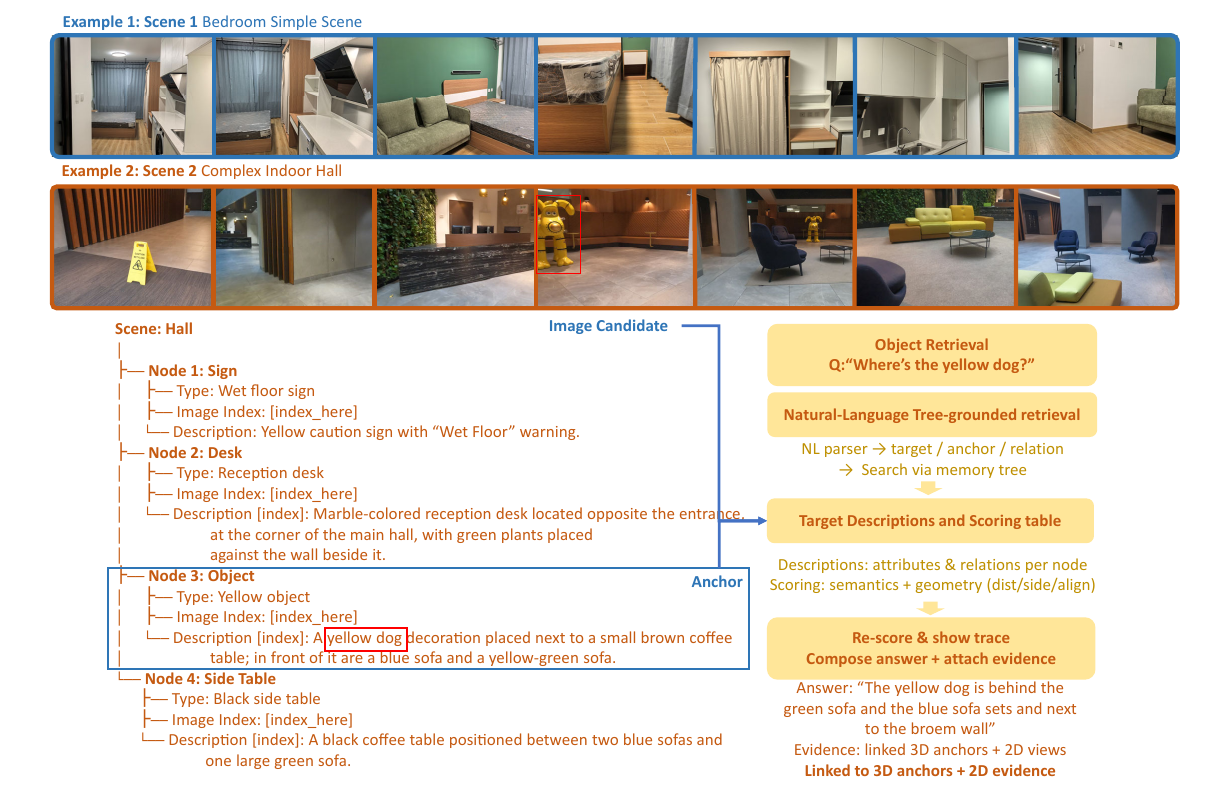}
    \caption{High-level SpatialMem pipeline. Left to right: egocentric RGB input, geometry and metric alignment, anchor and object construction, memory tree and query execution using geometry and semantic reasoning.}
    \label{fig:supp_pipeline}
\end{figure*}

\begin{figure*}[t]
    \centering
    \includegraphics[width=\linewidth]{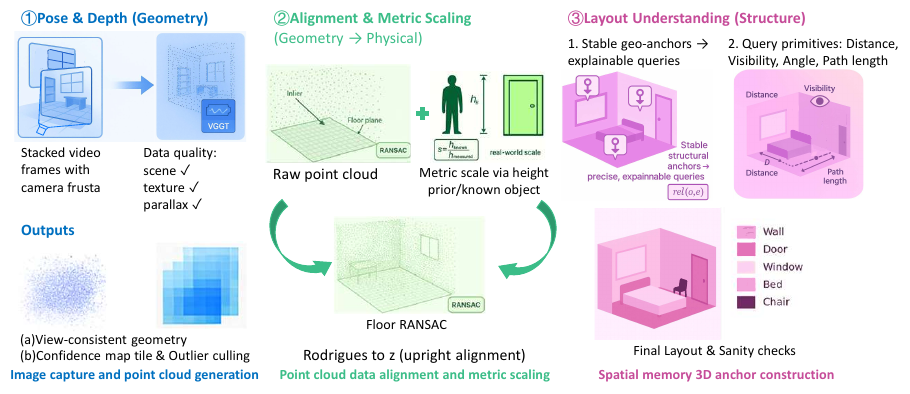}
    \caption{The geometry back-end and metric alignment. We use a feed-forward RGB geometry module and align it to an upright, metric frame.}
    \label{fig:stage1}
\end{figure*}

\begin{figure*}[t]
    \centering
    \includegraphics[width=\linewidth]{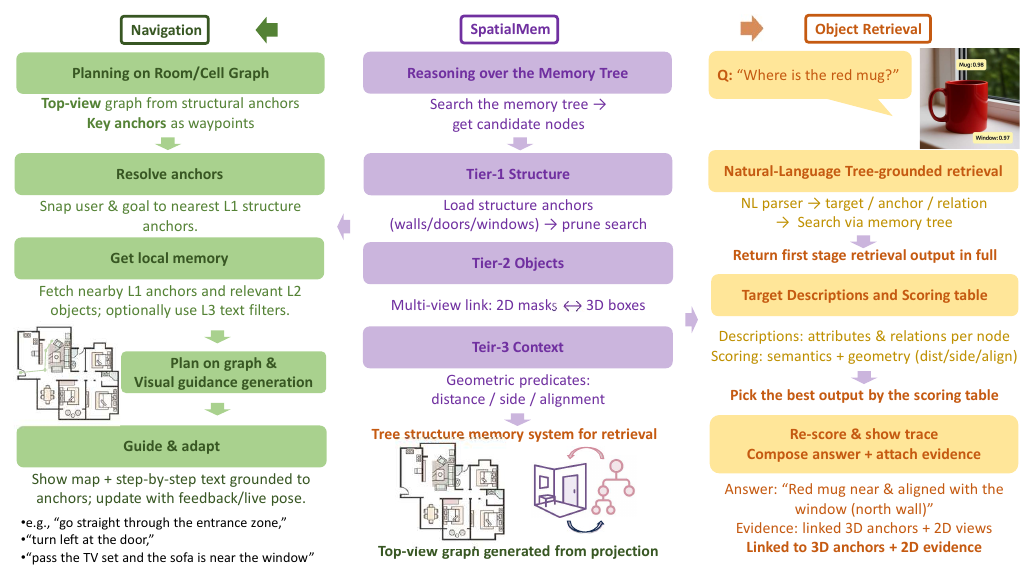}
    \caption{The SpatialMem pipeline overview. The memory tree (centre) orders structural anchors, objects and the contextual relations. On the left side: the offline guidance can be executed with the help of the anchor graph to establish user and goal locations and provide step-by-step graph guidance. On the right side: object retrieval processes a query in language, searches the tree, and returns an answer with 3D anchors and linked 2D evidence.}
    \label{fig:two_applications}
\end{figure*}

\section{Implementation Details and Shared Settings}
\label{sec:supp-impl}

This part will give a summary of the major implementation options that were employed in the pipeline. The same settings apply to all scenes unless specified otherwise and are not adjusted individually to any scene.

\begin{table}[t]
\centering
\caption{Shared implementation settings used in SpatialMem.}
\label{tab:supp-settings}
\scriptsize
\setlength{\tabcolsep}{4pt}
\renewcommand{\arraystretch}{1.05}
\resizebox{\columnwidth}{!}{
\begin{tabular}{lll}
\toprule
\textbf{Component} & \textbf{Setting} & \textbf{Value / rule} \\
\midrule
Frame sampling & video sampling stride & 10 \\
Floor fitting & RANSAC inlier threshold & 2--5\,cm \\
Metric scaling & height prior & fixed $2.5$\,m ceiling prior (no per-scene metadata) \\
3D box fitting & box type & axis-aligned bounding box (AABB) \\
Multi-view object association & grouping rule & center-distance matching \\
Duplicate suppression & merge range & approximately 25\,cm \\
Relative relations & proximity thresholds & $\tau_{\text{near}}=1.2$\,m, $\tau_{\text{attach}}=0.6$\,m, on-tol.\ $0.25$\,m \\
Guidance success & distance threshold $d_{\mathrm{thr}}$ & 1\,m \\
Relation consolidation & vote weight & anchor confidence \\
\bottomrule
\end{tabular}}
\end{table}

\paragraph{Shared settings across scenes.}
Frame sampling uses a stride of 10. To align with metrics, we use the floor plane with a RANSAC inlier threshold of between $2$--$5$\,cm. The metric scale is restored using a fixed $2.5$\,m ceiling-height prior applied identically to all scenes. No per-scene ceiling height, depth, pose, or intrinsic of any dataset is included in the construction path; a reconstruction whose implied scale factor is outside $[0.05,20]$ is discarded. The earlier entry in this table referred to the pilot captures and is superseded. We represent objects by fitting axis-aligned 3D boxes into the upright frame. Multi-view association is performed through centre-distance matching, and adjacent duplicates are filtered in a range of about $25$\,cm. We apply the thresholds indicated above to relation predicates like near and related spatial decisions; cases that are closer than $0.15$\,m to a threshold are excluded from the evaluation rather than adjudicated manually.

\paragraph{Manual verification for structural consistency.}
To construct anchors and merge structures, we use the principle of geometric consistency between views alone and do not have any human in the construction loop in the reported $287$-scene and $12$-scene experiments. In practical terms, planes and openings that are candidates are only retained if they continue to be visually and geometrically constant in several views in the aligned frame. This is to prevent the over-dependence on the scene-based numeric levels of anchor stability and provide a stable final anchor group based on the reconstructed layout.

\section{Dataset and Capture Details}
\label{sec:supp-dataset}

The capture setup and the pilot capture setup (two egocentric scenes and one public Replica room) will be summarized. All the data is based on the documentation of the project.

\subsection*{Capture Setup}

\begin{itemize}
  \item \textbf{Camera device:}
    \begin{itemize}
      \item Lightweight pin-style wearable camera, worn for hands-free egocentric recording.
      \item Compact form factor suitable for continuous indoor navigation.
    \end{itemize}

  \item \textbf{Capture protocol:}
    \begin{itemize}
      \item Natural egocentric walk-throughs with forward-facing motion.
      \item Single continuous pass per scene; indoor-only; no additional constraints specified.
    \end{itemize}
\end{itemize}

\subsection*{Scene Characteristics}

\begin{itemize}
  \item \textbf{HT (real egocentric hotel room):} a single furnished room; $6$ combined 3D items, $6$ walls / $2$ windows / $1$ door. Egocentric RGB recorded in a relaxed manner; the most descriptive per object annotation of all of the three scenes.

  \item \textbf{Apt (real multi-room apartment):} a larger multi-room layout; $14$ combined 3D objects across $46$ wall segments, $2$ doors, $2$ windows. The object-rich, multi-room setting is provided.

  \item \textbf{Replica (a public room):} a standardised reference of a public indoor scene; objects of the layout and dense images.
\end{itemize}

\section{Algorithmic Details for Memory Construction}
\label{sec:supp-algorithm}

The section will also add some additional operational information on anchor construction, object lifting and relation consolidation. It is intended that this makes it simpler to follow and reproduce the construction process without making the implementation itself too large.

\subsection{Anchor Construction and Structural Consolidation}
The construction of level-1 anchors is made using stable structural components including walls, doors and windows after geometric recovery and metric alignment.

\paragraph{Per-frame anchor proposals.}
The layout module generates structural candidates in image space per each sampled frame. The recovered camera pose and the associated scene geometry are used to lift these candidates into the aligned metric frame. All the candidates are in the form of a plane or box with a category label and a confidence level.

\paragraph{Cross-view verification.}
A candidate is considered a stable anchor only if it maintains geometric consistency in several views of the upright frame. Practically speaking, we determine whether the candidate has the same position, orientation and extent under repeated observation at close-by viewpoints. The cross-view recurrence and merge rules listed above will remove candidates with unstable geometry, inconsistent extent, or duplicate structure automatically. Manual curation was not used in any of the reported experiments; it was applied to the pilot captures only.

\paragraph{Anchor merging.}
Once the verification process is complete, structurally similar candidates of the same type are combined into a single anchor. Final anchor parameters are derived through refitting to the pooled geometry in the aligned frame. This merging phase is made conservative: in case two candidates are visually near but represent different physical structures, they remain unmerged; in case they are consistently seen as the same structure at various perspectives, they are merged and saved as one fixed anchor in the tree.

\paragraph{Anchor confidence.}
Every last anchor will save a confidence score that has been calculated through repetitive observation and structural uniformity. The score will be subsequently applied in relation consolidations whereby evidence supported by anchors is considered of greater value than a solitary view.

\subsection{Object Lifting from 2D Detections to 3D Boxes}
Level-2 object nodes are achieved through the lifting of open-vocabulary detections and masks out of the image space and into the metric scene frame.

\paragraph{Back-projection.}
We take each 2D detection/segmentation mask and cast the chosen image region into 3D with the recovered geometry and the aligned camera pose. This results in a collection of 3D points related to the object hypothesis in the metric frame.

\paragraph{Axis-aligned box fitting.}
In the upright scene frame, we use axis-aligned bounding boxes (AABBs) to represent each object. We then fit a tight box after the back-projection to cover the legitimate 3D points and discard outlier points that are clearly due to imperfect masks or background leaking. The final box contains its center, size, semantic label, and confidence.

\paragraph{Multi-view association.}
Object hypotheses based on various frames are related through center-distance matching in 3D. When two object boxes have compatible semantics and lie within about $25$\,cm of each other, they are considered to be the same physical object and combined into a single candidate. This is a straightforward association rule which can be applied in our indoor scenes since objects are already depicted in the same aligned metric frame.

\paragraph{Duplicate suppression.}
Following cross-view association, all close-by duplicates are deleted by saving the most stable combined box and ignoring duplicate local copies. It prevents the repetition of the same object in the memory tree because it is observed repeatedly by the nearby views.

\subsection{Lateral Relation Consolidation Across Views}
The first step in left-right-front-behind relations is collecting them locally based on individual perspectives and then combining them into a global memory.

\paragraph{Per-view relation evidence.}
In the case of every object-anchor or object-object pair that is relevant, we first note the local directional correlation that we observe in the given frame. The local tags are view-dependent and represent the egocentric view at that instant.

\paragraph{Global consolidation in the aligned frame.}
To reconcile local relation evidence in the upright metric frame, we use aligned anchor geometry. We do not treat all frames as being equal, but instead weight the contributions of individual observations on anchor confidence, such that relationships based on more stable structural anchors have a greater influence on the ultimate decision.

\paragraph{Final stored relation.}
The last relation label is chosen based on the weighed evidence gathered on views. In case of weak or unstable observations, the system maintains a local descriptive relationship rather than imposing a strong but possibly incorrect overall relationship. It makes the stored memory more conservative in hard situations.

\begin{figure*}[t]
    \centering
    \includegraphics[width=\linewidth]{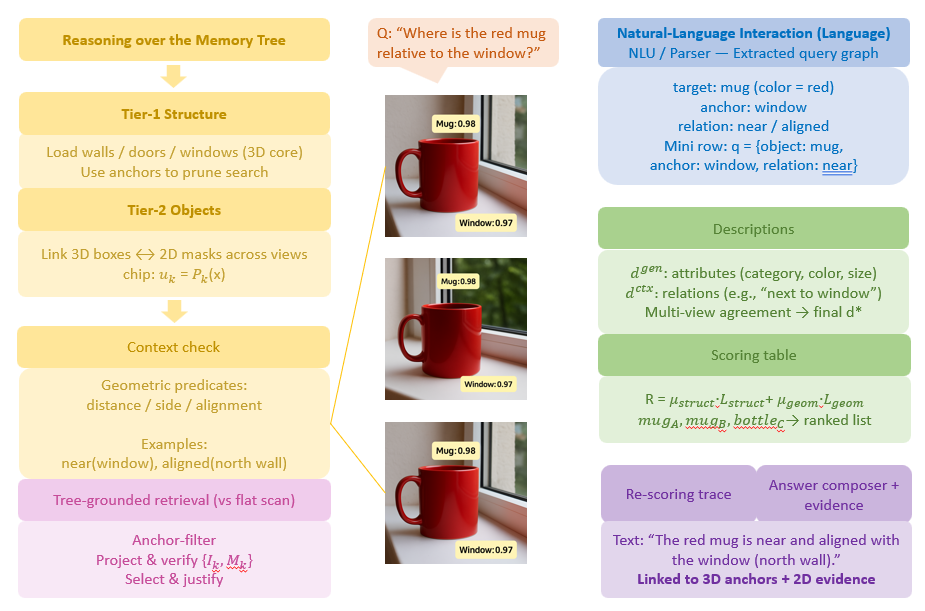}
    \caption{Relational query traversal using the memory tree: The parsed query graph is solved against Tier-1 anchors and Tier-2 objects, candidates are filtered by geometric predicates and description agreement and the result is constructed with connected 3D anchors and 2D evidence.}
    \label{fig:supp_reasoning}
\end{figure*}
\subsection{Memory-Tree Assembly}
After anchor and object construction, we assemble the final hierarchical memory.

\paragraph{Level-1: structural anchors.}
Solid walls, doors and windows and other architectural features are stored as Level-1 nodes. These anchors give the spatial framework of the scene.

\paragraph{Level-2: lifted objects.}
The lifted 3D objects are coupled to their nearest or support anchor based on metric relationships in the aligned space. It generates a grounded object layer which remains attached to explicit scene structure.

\paragraph{Level-3: text descriptions.}
Every object node contains language descriptions in two layers: short general description and more contextual one containing nearest anchors, relations or characterizing scene details. They are then used to answer queries and retrieve information.

\section{Expanded Metrics and Error Analysis}
\label{sec:supp-metrics}

The given section builds on the metric definitions given in Sec.~\ref{sec:metrics} by giving complete formulas and operational information so that they can be reproduced.

\subsection{Evaluation Protocol}
\label{sec:supp-protocol}
The stated experiments measure the following; they replace the formulas in what follows, which are kept as a reference. (i) Relation verification accuracy: balanced yes/no probing on viewpoint-invariant relations (near/on/above/below) with hard negatives, chance=0.5. (ii) Cross-view consistency: the fraction of episodes whose answers are the same in three disjoint temporal frame segments (with no canonical answer presupposed). (iii) Retrieval: $\mathrm{SR}_{\text{obj}}$ (exact label), $\mathrm{Acc}_{\text{parent}}$ and $\mathrm{Acc}_{\text{path}}$ (exact anchor / root$\rightarrow$leaf path). (iv) Description: token-level colour recall with respect to confirmed attributes. (v) Efficiency: wall-clock per-query latency. All measurements are done using $95\%$ bootstrap confidence intervals on the $174$ human-verified episodes of the two egocentric captures; the Replica room is discussed separately below.

\paragraph{Why single-label relation accuracy is not used head-to-head.} A couple usually fulfills a number of relationships simultaneously (a bed may be both close to, to the left of, and even in front of a curtain) and lateral relationships depend on viewpoints. Scoring one answer against a single designated relation penalises legitimate alternatives and confuses frame conventions, hence we use verification + consistency.

\paragraph{Navigation.} The report of navigation is qualitative (path-based guidance on the prebuilt plan). The navigation metrics that follow (SR$_{\text{nav}}$, StepComp, SPL, NE) are kept on record, but they will not be used as quantitative measures in this version since it focuses on offline memory instead of closed-loop control.

\begin{table}[h]
\centering
\caption{Per-scene results of SpatialMem (human-verified GT, $95\%$ CI). Relation / parent / path have been limited by reconstruction (answered by geometry); VLM comparison has been grouped into the main paper because it contains small per-scene samples.}
\label{tab:supp-perscene}
\scriptsize
\setlength{\tabcolsep}{5pt}
\begin{tabular}{lcccc}
\toprule
Scene & $\mathrm{Acc}_{\text{rel}}$ & $\mathrm{SR}_{\text{obj}}$ & $\mathrm{Acc}_{\text{parent}}$ & $\mathrm{Acc}_{\text{path}}$ \\
\midrule
HT (hotel room)   & 0.96 & 1.00 & 0.83 & 0.83 \\
Apt (multi-room)  & 0.99 & 1.00 & 0.79 & 0.79 \\
Replica (public)  & 1.00 & 1.00 & 1.00 & 1.00 \\
\bottomrule
\end{tabular}
\end{table}

\subsection{Notation}

Let $\mathcal{D}$ be the episode set. In the case of episode $i\in\mathcal{D}$, let the goal location be $g_i$, the end-user location be $\hat g_i$, the shortest path length be $L_i^\star$ and the executed path length be $L_i$. Let $\mathbf{1}[\cdot]$ be the indicator function. To navigate, we employ: a success distance limit $d_{\mathrm{thr}}$ (e.g., $1$\,m), proximity limits $\tau_{\text{near}},\tau_{\text{attach}}$, and an angular tolerance $\theta_{\text{rel}}$ used in left/right/front/back decisions. We provide mean values across scenes with $95\%$ bootstrap confidence intervals.

\subsection{Ground-Truth Construction and Quality Control}
\label{sec:supp-gt}

The ground truth of relations, guidance episodes, and retrieval paths is constructed out of the aligned memory as well as manual checks of the original egocentric videos.

\paragraph{Relation labels.}
In Task 1, all relation queries are compared with the aligned 3D layout and the related visual data in the video. The end label can be given only if the relative positioning is evident in the metric frame. When there are boundary cases close to decision boundaries, we do a slight manual audit to ensure that we do not penalize examples with visual ambiguity.

\paragraph{Guidance labels.}
In Task 2, start points, target points and decision steps are defined on the verified topology of the scene. The fulfilment of a step is compared with the planned structural course, the visible landmarks, and the position of the end goal in the aligned scene.

\paragraph{Hierarchical retrieval labels.}
In Task 3, the ground-truth path of a target object is determined from the root to the object node via its structural context. The object box, its nearest anchor and the corresponding descriptive evidence stored in memory are used to verify this path.

\paragraph{Quality control.}
The three tasks are all based on the same idea that a label will be accepted only when the metric reconstruction and the original visual evidence coincide. It means that the assessment is focused on the real structure of the scene rather than on the assessment made through words only.

\subsection{Task 1: Basic Relative Position Evaluation}
\label{sec:supp-rel-metrics}

This task evaluates whether the memory recovers basic spatial relations between entities
\[
\mathrm{rel} \in \{\text{left},\text{right},\text{front},\text{behind},\text{near},\text{on}\}.
\]
Every request identifies two targets, and the system responds with a statement of the form ``$A$ is $\mathrm{rel}$ of $B$.''

Consider $\mathcal{Q}$ to be the set of queries with prediction relation $\widehat{\mathrm{rel}}_q$ and ground-truth relation $\mathrm{rel}^\star_q$. The primary measure that is used is the Relation Accuracy:
\begin{equation}
\mathrm{Acc}_{\text{rel}}
\;=\;
\frac{1}{|\mathcal{Q}|}
\sum_{q\in\mathcal{Q}}
\mathbf{1}\!\left[
\, \widehat{\mathrm{rel}}_{q} = \mathrm{rel}^\star_{q} \,
\right].
\end{equation}

We also provide accuracy by anchors. Given any Level-1 structural anchor s (e.g., a specific wall, door or window) we calculate $\mathrm{Acc}_{\text{rel}}(s)$ on queries referring to only s and report a macro-average across the anchors:
\begin{equation}
\mathrm{Acc}_{\text{rel}}^{\text{macro}}
\;=\;
\frac{1}{|\mathcal{S}|}
\sum_{s\in\mathcal{S}}
\mathrm{Acc}_{\text{rel}}(s).
\end{equation}

\subsubsection*{Operational definitions (relative relations)}

The orientation towards left/right/front/behind is calculated in the allocentric, upright frame. With a structural anchor of outward normal and a target object box, we determine the side relationship based on the signed angle between the reference direction of the anchor and the vector between the anchor and the object; predictions that fall within $\theta_{\text{rel}}$ of the annotated side are considered correct.

Considering near/attached predicates, we define the 3D box of object o as $B_o$ and the plane/box of anchor e as $(\pi_e, B_e)$. We have:
\begin{itemize}
  \item \texttt{attached} if
  $\min_{x\in B_o} \operatorname{dist}(x,\pi_e) < \tau_{\text{attach}}$,
  \item \texttt{near} if
  $\min_{x\in B_o} \operatorname{dist}(x,B_e) < \tau_{\text{near}}$,
\end{itemize}
where the thresholds are selected empirically and common to all scenes. The generator also eliminates episodes with a geometry that is closer than $0.15$\,m to a predicate threshold, and instead of spot checks, the human audit of Sec.~\ref{sec:supp-protocol} records a verdict on each sampled episode.

\paragraph{Practical thresholds.}
The given range is common between the scenes and provides a constant regulation of the relation between indoor objects and anchors without repeating the process in each scene.

\subsection{Task 2: Navigation Metrics (reference only)}
\noindent The definitions below are retained for reference only; path guidance is reported qualitatively, and no number in the paper is computed with these metrics.

\label{sec:supp-nav-metrics}

We break down navigation performance into (i) success rate, (ii) step-weighted completion, (iii) path efficiency and (iv) error.

\paragraph{Offline guidance setting.}
The navigation task discussed in this paper is assessed as a memory-based offline guidance of a previously constructed memory instead of an online embodied control. With a starting point and a target, the system provides step-by-step instructions on how to navigate a validated structure graph based on the reconstructed scene.

\paragraph{Decision-point graph construction.}
The graph is constructed based upon manually checked structural transitions, including the entry of rooms, openings and significant turning points. The edges are traversable connections in the aligned layout. On this graph, shortest paths are calculated and taken as the reference path lengths $L_i^\star$. Path length executed $L_i$ is measured on the associated predicted guidance sequence.

\subsubsection*{Navigation success rate}

If the final predicted goal is inside distance $d_{\mathrm{thr}}$ of the actual target, then an episode can be considered to have been successful:
\begin{equation}
S_i^{\text{nav}}
=
\mathbf{1}\!\left[
\bigl\lVert \hat{g}_i - g_i \bigr\rVert_2
\le
d_{\mathrm{thr}}
\right].
\end{equation}
The navigation success rate is
\begin{equation}
\mathrm{SR}_{\text{nav}}
=
\frac{1}{|\mathcal{D}|}
\sum_{i\in\mathcal{D}}
S_i^{\text{nav}} \, .
\end{equation}

\subsubsection*{Step-weighted navigation completion}

All navigation descriptions have $T_i$ ordered steps, which are completed using score $c_{i,t}\in[0,1]$ and non-negative weights $w_{i,t}$ (subsequent steps may be given more weight). Per-episode completion is
\begin{equation}
\mathrm{StepComp}_i
=
\frac{\sum_{t=1}^{T_i} w_{i,t}\, c_{i,t}}{\sum_{t=1}^{T_i} w_{i,t}},
\end{equation}
and we report the average
\begin{equation}
\overline{\mathrm{StepComp}}
=
\frac{1}{|\mathcal{D}|}
\sum_{i\in\mathcal{D}}
\mathrm{StepComp}_i .
\end{equation}

\paragraph*{Operational definition (step completion).}
In textual steps, $c_{i,t}=1$ if the step's verifiable condition (landmark reached/visible, turn executed) is satisfied by logs and geometry; in case of partial satisfaction (e.g., landmark reached but additional detour was done), graded values in (0,1) are used.

\subsubsection*{Path efficiency (SPL)}

We use the standard Success-weighted by Path Length (SPL):
\begin{equation}
\mathrm{SPL}
=
\frac{1}{|\mathcal{D}|}
\sum_{i\in\mathcal{D}}
S_i^{\text{nav}}
\cdot
\frac{L_i^\star}{\max(L_i,\, L_i^\star)} \, ,
\end{equation}
where $L_i^\star$ is the shortest-path length to the goal and $L_i$ is the length of the executed path.

\subsubsection*{Navigation error (NE)}

Navigation error is the mean final distance to the goal over failed episodes:
\begin{equation}
\mathrm{NE}
=
\frac{1}
{\left|\{\, i \in \mathcal{D} : S_i^{\text{nav}}=0 \,\}\right|}
\sum_{i \in \mathcal{D} : S_i^{\text{nav}}=0}
\bigl\lVert \hat{g}_i - g_i \bigr\rVert_2 .
\end{equation}
The distance thresholds, angular tolerances, and the rules of tie breaking (e.g., in case of several near-optimal paths) are defined to be consistent with the main text.

\paragraph{Success threshold.}
Our navigation success threshold is $d_{\mathrm{thr}} = 1\,\mathrm{m}$. The same choice is made in all the scenes and it is suitable to be used as a decision-point guidance in an indoor setting.

\subsection{Task 3: Object Retrieval Metrics}
\label{sec:supp-obj-metrics}

The object retrieval episodes consist of finding a target object $o_i$ and verifying it (e.g., describing it or pointing at it).

\subsubsection*{Retrieval success}

Let
\[
R_i
=
\mathbf{1}\!\left[\text{correct object $o_i$ confirmed}\right].
\]
The object retrieval success rate is
\begin{equation}
\mathrm{SR}_{\text{obj}}
=
\frac{1}{|\mathcal{D}|}
\sum_{i\in\mathcal{D}} R_i \, .
\end{equation}
When retrieval depends on navigation, an episode is considered fully successful only when the goal is reached and the object is confirmed; we also provide the two components separately in the main tables.

\subsubsection*{Hierarchical (tree) retrieval accuracy}

Every retrieved object node $\hat{v}_i$ has a parent chain from the root to $\operatorname{parent}(\hat{v}_i)$. Set $v_i$ as the ground-truth node.

\emph{Parent accuracy} measures whether the immediate parent (e.g., container, room, or anchor) is correct:
\begin{equation}
\mathrm{Acc}_{\text{parent}}
=
\frac{1}{|\mathcal{D}|}
\sum_{i\in\mathcal{D}}
\mathbf{1}\!\left[
\operatorname{parent}(\hat{v}_i)
=
\operatorname{parent}(v_i)
\right].
\end{equation}

\emph{Path accuracy} requires the entire root$\rightarrow$leaf path to match:
\begin{equation}
\mathrm{Acc}_{\text{path}}
=
\frac{1}{|\mathcal{D}|}
\sum_{i\in\mathcal{D}}
\mathbf{1}\!\left[
\operatorname{path}(\hat{v}_i)
=
\operatorname{path}(v_i)
\right].
\end{equation}

\paragraph{Hierarchical verification.}
In hierarchical retrieval, the correctness is compared with the complete rooted memory instead of comparing it with just the end object name. It is a more stringent measure in cluttered environments because the object that has been retrieved also has to be anchored to the appropriate structure.

\subsection{Description Quality Metrics}
\label{sec:supp-desc-metrics}

We evaluate the quality of language descriptions paired with memory.

\begin{itemize}
  \item \textbf{Color Accuracy (Color Acc):} the proportion of episodes in which the described color properties (e.g., ``red mug'', ``blue sofa'') are consistent with the ground-truth labels.
  \item \textbf{Loc@parent:} proportion of episodes in which the location phrase accurately describes the parent container or anchor in the tree (e.g., on the side table next to the green sofa).
  \item \textbf{Loc@path:} a more strict version that requires the entire root$\rightarrow$leaf path implied by the description to be the same as ground-truth path (scene, room/anchor, container, object).
\end{itemize}

Such metrics are calculated based on human-labeled ground truth and on the text outputs matched with the given memory nodes.

\paragraph{Description grounding.}
The description is considered together with memory node to which it has been connected. It implies that a smooth sentence cannot be alone but the stated properties and location clues should also correspond to the physical object and the hierarchy in the tree.

\subsection{Instruction Quality and Error Analysis}
\label{sec:supp-error-metrics}

The analysis of the quality of step-by-step instructions and the types of errors that can be made is performed independently of the raw success rates.

\subsubsection*{Category proportions}

Let $E_{\mathrm{P}},E_{\mathrm{R}},E_{\mathrm{M}}$ be counts of Planning, Reasoning, and Perception/Memory errors over all labeled errors (an episode may have multiple tags). We report their proportions:
\begin{align}
\mathrm{Err}_{\mathrm{P}}
&=
\frac{E_{\mathrm{P}}}{E_{\mathrm{P}}+E_{\mathrm{R}}+E_{\mathrm{M}}}, \\
\mathrm{Err}_{\mathrm{R}}
&=
\frac{E_{\mathrm{R}}}{E_{\mathrm{P}}+E_{\mathrm{R}}+E_{\mathrm{M}}}, \\
\mathrm{Err}_{\mathrm{M}}
&=
\frac{E_{\mathrm{M}}}{E_{\mathrm{P}}+E_{\mathrm{R}}+E_{\mathrm{M}}}.
\end{align}

\subsubsection*{Subtype breakdown}

For each top-level type $X\in\{\mathrm{P},\mathrm{R},\mathrm{M}\}$ and subtype “sub” with count $E_{X,\text{sub}}$, we define:
\begin{align}
\mathrm{Err}_{X,\text{sub}\,|\,X}
&=
\frac{E_{X,\text{sub}}}{E_X},
\\
\mathrm{Err}_{X,\text{sub}}^{\text{global}}
&=
\frac{E_{X,\text{sub}}}{E_{\mathrm{P}}+E_{\mathrm{R}}+E_{\mathrm{M}}}.
\end{align}

The subtypes are:
\begin{itemize}
  \item \emph{Planning} --- wrong goal/room, missed step, wrong order, inefficient detour.
  \item \emph{Reasoning} --- spatial relation mistake, logical inconsistency, insufficient exploration.
  \item \emph{Perception/Memory} --- hallucinated object/landmark, wrong recognition, mislocalization.
\end{itemize}

\paragraph{Interpretation of error tags.}
The tags are intended to distinguish between the failure being primarily due to path construction, relational reasoning, or memory grounding. In a complex indoor scene, an episode could have several sources of failure, so the error analysis is done at the tag level instead of requiring only one label per episode.

\section{Queryable-Index Footprint Accounting}
\label{sec:supp-footprint}
The $14.3$\,KB median per-scene number quoted in the main paper is precisely what the query scan reads, i.e., the layout file (walls, doors, and windows with plane and box parameters), the object nodes (labels, oriented boxes, view counts, and first-sighting frames), the relation edges, and the two description layers as text. The index stores an average of roughly $240$ bytes per node, with the text included, because the median memory stores some $60$ nodes. Not included are the aligned point cloud ($2.6$\,MB per scene), reconstruction intermediates ($22.5$\,MB per scene, discardable), and the source frames ($1.9$\,MB per scene), and consequently also the evidence frames that retrieval-mode answering presents, which are read from the frame store only after the index has selected them. It should be noted that $14.3$\,KB is what it costs to decide what to look at, not what it costs to store what has been looked at.

\section{Runtime Breakdown}
\label{sec:supp-runtime}

In our report we have more timing statistics on the SpatialMem pipeline such as the end-to-end preprocessing time and the per-task testing-time information that is mentioned in the main paper.

\paragraph{Offline preprocessing.}
The SpatialMem needs an initial preprocessing step that will transform the egocentric RGB stream into a metric 3D memory. The total pipeline time is:

\begin{itemize}
  \item \textbf{Point-cloud generation and alignment:} Median $\sim\!100$\,s for feed-forward geometry, fusion and gravity alignment on the benchmark videos.

  \item \textbf{3D bounding-box extraction and automatic annotation:} Median $16$\,s for the layout model and $18$\,s (p90 $278$\,s) for lifting 2D detections and masks to 3D and describing them; the long tail is scenes with large numbers of detected instances, as the description model is run per object.

  \item \textbf{Memory-tree construction and vector-database generation:} Median $7$\,s. The process of assembling the tree and the queryable index is arithmetic on already-extracted boxes and not a learned step.
\end{itemize}

The sum of the stage medians is the $118$\,s per-scene median (p90 $381$\,s) as presented in the main text, measured across all $287$ scenes on one GPU.

\paragraph{Task-level inference time.}
The index scan behind the low-latency claim is measured directly on the
OpenEQA retrieval path, with no language model in the loop. Over the $953$
questions of the $79$ ScanNet scenes, selecting the evidence frames for a
question takes a median $0.20$\,ms per query (p95 $2.7$\,ms) on one CPU core,
and loading a scene's queryable index takes a median $1.1$\,ms once per scene.
The cost is flat in scene size over the observed range of $19$--$172$ nodes
($0.18$\,ms median on the smaller half of the scenes against $0.19$\,ms on the
larger half). These timings cover the memory's selection path only; the
answering model's inference is a separate cost shared by every method.

\paragraph{Practical interpretation.}
The major computational expense is in the one-off offline construction step. After the metric memory has been constructed, subsequent query answering is relatively inexpensive, and largely depends upon tree traversal, geometric filtering, and language generation given a fixed memory. The design allows using the system in large and multi-room scenes because the heavy geometry step is executed once and subsequent queries are answered using the saved structure.

\paragraph{Memory efficiency.}
SpatialMem holds a specific scene memory that is explicitly stored and that increases with scene complexity, but still significantly less than the runtime memory footprint of local end-to-end VLM inference. The operational footprint is the $14.3$\,KB median searchable index measured across the $287$ scenes ($4.07$\,MB overall).

\section{Additional Setup Details}
\label{sec:supp-r2}

\paragraph{Suite scene identifiers.}
The $12$ ScanNet scenes that are included are \texttt{scene0207\_02}, \texttt{scene0645\_00}, \texttt{scene0591\_01}, \texttt{scene0608\_00}, \texttt{scene0643\_00}, \texttt{scene0697\_01}, \texttt{scene0353\_00}, \texttt{scene0580\_01}, \texttt{scene0653\_00}, \texttt{scene0426\_00}, \texttt{scene0593\_00} and \texttt{scene0231\_00}, and chosen because they have both VSI-Bench coverage and all official instance annotations.

\paragraph{Ingestion settings.}
Egocentric capture is recorded at 3840x2160 (photo) and 2016x1512@30 fps, scaled to 1024x768-1920x1080 on ingestion; the geometry back-end operates at 640x480-1024x768 and frames are sampled every $1$--$2$\,s. On the publicly available benchmark videos we use $40$--$96$ frames per scene, sampled uniformly.

\paragraph{Frame-based baseline protocol.}
Every model is given $8$ evenly sampled frames of each scene, which are JPEG encoded with $512$ pixels on the longer edge and the same question text as ours. The relation probe questions are posed in the form of the same balanced yes/no question; the retrieval question is in the form of presence and a one sentence description of the location. Every one of the four models parsed cleanly in all relation episodes ($0$/$573$).

\paragraph{Scoring-time registration.}
We are estimating the $4$-DoF similarity (scale, yaw, translation) between our reconstruction frame and the ScanNet axis-aligned frame per scene. The candidate transforms are produced by matching the two room footprints based on the four right-angle yaw hypothesis; the candidates are evaluated as to how many uniquely labelled object pairs they bring to less than 0.6m, and the highest scoring candidate is further optimized over its inliers. All the $12$ scenes are registered with a median of residuals of $0.44$\,m. The transform is used in the comparison of predicted centres and ground truth only.

\paragraph{Memory footprint accounting.}
Queryable index is composed of the layout file, the object nodes and their description (median $14.3$\,KB per scene). Aligned point cloud (median $2.6$\,MB) and reconstruction intermediates (median $22.5$\,MB) are reported separately and none of them are read at query time.

\section{Additional Experiments and Full Metric Definitions}
This section collects experiments summarized in a single sentence of the main text, together with the full metric definitions the main text abbreviates.

\subsection{Protocol Details Moved from the Setup Section}
The metric definitions that the setup section now summarises in a table are reproduced below in full and unchanged, together with the notation used by the guidance formulas.

Let $\mathcal{D}$ be the set of episodes. Episode $i\in\mathcal{D}$ will indicate goal $g_i$, the final user location $\hat g_i$, shortest path $L_i^\star$, executed path $L_i$, and indicator $\mathbf{1}[\cdot]$. In this case, $L_i^\star$ represents the shortest possible length between the starting point and the goal on an identical 2D floor-projected decision-point graph, whereas $L_i$ is the result of projecting the estimated waypoint sequence into the same graph. Navigation takes into account a distance limit $d_{\mathrm{thr}}$, proximity limits $\tau_{\text{near}},\tau_{\text{attach}}$, and angle limits $\theta_{\text{rel}}$.

\noindent\textbf{Evaluation frame.} Relations such as left/right/front/behind are evaluated in an allocentric frame with a fixed angular tolerance; \texttt{near}/\texttt{attached} use distance thresholds to the plane or box anchor. We macro-average over anchors and include a light audit for ambiguous boundary cases.

 The two forms are reported for the self-captured scenes in Sec.~\ref{sec:supp-egocentric}. The annotation-derived suite has probes that are balanced yes/no instead of being open-ended, and thus the main table reports accuracy across all episodes, the matched episode set, and per predicate family (proximity versus vertical) rather than per anchor type; hierarchy is where the old per-anchor-type breakdown is now located.

There are two changes. First, the ground truth does not come out of our own reconstruction anymore: it is supplied by the official ScanNet instance annotations on the suite, and by the benchmark's released answers on VSI-Bench. Secondly, since our answerer does not give any answer when an entity asked about is not in memory but the scene-graph baseline will always return the nearest match, we provide accuracy both over all episodes and over those that can be resolved. The frame-based protocol is provided in Sec.~\ref{sec:supp-impl}.

 The compound criterion applies only to the guidance demonstration kept in Sec.~\ref{sec:supp-nav-retained}; on the suite, retrieval success is the $0.5$\,m centre criterion defined below, and we also state a relaxed $1.0$\,m radius.

 $Acc_{\text{path}}$ demands that our node indices should match the annotated ones, a requirement no cross-system comparison can satisfy, therefore it is reported only on the self-captured scenes in Sec.~\ref{sec:supp-egocentric}.

 We also report Color Acc based on human color labels on the suite because ScanNet does not label colors; the two location-phrase metrics have the same node-correspondence issue discussed earlier and are reported solely in Sec.~\ref{sec:supp-egocentric}.

\noindent\textbf{Metrics summary.} The main tables report, on VSI-Bench, the benchmark's eight task scores and their macro mean; and on the suite, relation verification accuracy (all episodes, matched episode set, and per predicate family), $\mathrm{SR}_{\text{obj}}$ at $0.5$ and $1.0$\,m, $\mathrm{Acc}_{\text{parent}}$ overall and per annotated parent type, and \textit{Color Acc}. The remaining metrics defined above --- $\mathrm{Acc}_{\text{rel}}$ and $\mathrm{Acc}_{\text{rel}}(s)$ in their open-ended form, $\mathrm{SR}_{\text{nav}}$, $\overline{\mathrm{StepComp}}$, $\mathrm{SPL}$, $\mathrm{NE}$, $\mathrm{Acc}_{\text{path}}$, \textit{Loc@parent} and \textit{Loc@path} --- are reported only for the self-captured scenes in Sec.~\ref{sec:supp-egocentric}, and no quantitative claim in the main paper rests on them.

The setup section gives the protocol in summary form; the full text (unchanged) is given below.

\noindent\textbf{Ground truth and query generation (suite).} The queries are created in a deterministic manner (seed $42$) based on the official annotations: relation probes are balanced yes/no questions in terms of viewpoint-invariant predicates (near, on, above, below) with strong negatives; retrieval queries identify a target by a label and a relational clue and are kept only when the target can be uniquely identified under a $\pm0.15$\,m robustness range. Episodes that fall within $0.15$\,m of a predicate boundary are eliminated as ambiguous, and the predicate boundaries correspond precisely to the answering code ($\tau_{\text{near}}=1.2$ m, $\tau_{\text{attach}}=0.6$ m, on-tolerance $0.25$\,m). A human audit considers $200$ relations ($35\%$), $124$ retrievals ($26\%$), and all $180$ description items, totaling $504$ judgments: five were unviewable and seven were judged incorrect, so generator precision is $492/499=98.6\%$, and those twelve episodes are omitted for every system, leaving the $573$ relation and $471$ retrieval episodes all tables report. In order to enhance reproducibility, we are releasing the suites, the constructed memories and the evaluation code, and the real-world scenes with their query sets and protocol.

Here are two changes. Firstly, the ground truth is no longer proposed from our own reconstruction: in the suite it is based on the official ScanNet instance annotations, and in VSI-Bench on the benchmark's released answers. Secondly, since our answerer does not respond when an entity being asked is not present in memory whereas a scene-graph baseline will always produce its nearest answer, the suite probes are balanced yes/no questions (chance $0.5$) and we provide accuracy across all episodes and on the ones that are solvable. The frame-based multimodal models (GPT-4o, Gemini-2.5-Flash, Qwen2.5-VL-72B, GPT-4o-mini) take uniform samples of eight frames of size $512$\,px with the same question text.

These definitions require one modification relative to an external baseline, which we apply to both systems in a similar manner: our anchor indices are not able to be aligned to ScanNet instance ids, and therefore a predicted parent is regarded as correct if its type is the same as the annotated parent type and our anchor is within $1.0$\,m of the annotated anchor centre. $\mathrm{SR}_{\text{obj}}$ maintains the $0.5$\,m centre criterion, which is also the baseline's own. Since our RGB-only memory exists in a self-estimated frame whereas the annotations exist in ScanNet's, we estimate a $4$-DoF similarity (scale, yaw, translation) per scene at scoring time only; it never becomes part of the answering path, and the resulting error is attributed to our system. The matched-input variant of the results section of the main paper is built from the official posed depth with the released axis-align matrix and does not require such a step.

\subsection{Resolved-Entity Split of the Metric Tasks}
The concern implied in the request is that a system score does not indicate the identity of the component that generated the answer. Each saved answer is associated with the handler that created it and mapping entity used, and thus, the two populations may be separated precisely.

\begin{table}[t]
\centering
\caption{The VSI-Bench metric tasks split by whether the queried entities resolved to memory objects ($95\%$ CI). Only room size is answered from stored geometry throughout; absolute distance mostly is not, and where it is, the geometry does not beat the estimate. That is why we do not rest the argument on that task.}
\label{tab:resolved}
\scriptsize
\setlength{\tabcolsep}{4pt}
\begin{tabular}{lccc}
\toprule
\textbf{Task} & \textbf{resolved} & \textbf{not resolved} & \textbf{\% resolved} \\
\midrule
Room size        & \textbf{41.4} [37.4,45.3] & --                & 100 \\
Counting         & 33.7 [30.0,37.7] & 0.0               & 74 \\
Object size      & 52.3 [48.9,55.9] & 51.9 [48.5,55.5]  & 48 \\
Absolute distance & 25.1 [20.8,29.3] & 28.6 [25.6,31.3] & 30 \\
\bottomrule
\end{tabular}
\end{table}

\subsection{Accuracy versus Video Duration}
A natural question is whether the framework applies to longer video sequences. Since the budget of frames is constant, a longer capture has a lower sampling rate, hence we divided the $287$ scenes into four duration quartiles. Accuracy does not decrease with length but increases, from $31.2$ $[27.1,35.9]$ in the shortest quartile ($22$--$56$\,s) to $40.4$ $[37.9,42.8]$ in the longest ($120$--$434$\,s), and entity resolution remains unchanged ($0.52$--$0.61$). A reasonable hypothesis is that a longer tour will revisit the same objects at more perspectives, and this supports the multi-view association on which the memory relies. We observe the confound that duration and source dataset are correlated, which we do not control for.

\subsection{Relation Verification on the Matched Episode Set}
We avoid responding to a queried entity not present in memory whereas none of the baselines does so, thus comparing our answerable-episode accuracy to the all-episode accuracy of a baseline would be comparing two different sets of questions. The main text therefore also reports each system on one matched set: the $390$ episodes that both of our memories resolve, the same denominator across all seven rows. The all-episode column, in which an abstention is considered an error, is provided alongside so that the cost of abstaining is visible.

\begin{table}[h]
\centering
\caption{Relation verification on the matched set (n=$390$; $95\%$ bootstrap CI) beside the all-episode value (n=$573$). The lead our matched-input memory holds on the matched set is nominal: its confidence interval overlaps the intervals of the scene graph and the most powerful frame-based models.}
\label{tab:matched}
\scriptsize
\setlength{\tabcolsep}{4pt}
\begin{tabular}{lcc}
\toprule
\textbf{System} & \textbf{matched} \tiny(390) & \textbf{all episodes} \tiny(573) \\
\midrule
GPT-4o                   & $0.615$ $[0.564,0.664]$ & $0.613$ \\
Gemini-2.5-Flash         & $0.623$ $[0.574,0.669]$ & $0.604$ \\
Qwen2.5-VL-72B           & $0.610$ $[0.559,0.659]$ & $0.586$ \\
GPT-4o-mini              & $0.549$ $[0.495,0.600]$ & $0.548$ \\
ConceptGraphs            & $0.623$ $[0.569,0.669]$ & $0.611$ \\
\midrule
\textbf{SpatialMem-RGBD} & $\mathbf{0.667}$ $[0.621,0.718]$ & $0.553$ \\
\textbf{SpatialMem}      & $0.610$ $[0.564,0.659]$ & $0.433$ \\
\bottomrule
\end{tabular}
\end{table}

\subsection{Per-Task Object-Layer Ablation and Entity Coverage}
The object-layer ablation is listed in the main text as a single row of Tab.~\ref{tab:vsi} and one overall figure. The breakdown by task is shown below, beside the percentage of questions whose queried entities all resolve to memory objects. This percentage is a question-level quantity and does not correspond to the entity-level coverage of $56.3\%$ mentioned in the main text. Taken together, they show that the pattern is not some random occurrence; the layer pays off on the tasks that ask what is in the room and costs a little where a coarse detection box replaces the layout's own extent. $\Delta$ is the difference between the two values displayed.

\begin{table}[h]
\centering
\caption{Object layer ablated on all $5{,}130$ VSI-Bench questions.}
\label{tab:supp-objlayer}
\scriptsize
\setlength{\tabcolsep}{3pt}
\resizebox{\columnwidth}{!}{
\begin{tabular}{lcccc}
\toprule
\textbf{Task} & \textbf{Full} & \textbf{w/o obj.\ nodes} & $\Delta$ & \textbf{quest.\ resolved} \\
\midrule
Object count       & 24.9 & 13.9 & $+11.0$ & 74\% \\
Absolute distance  & 27.5 & 28.9 & $-1.4$ & 30\% \\
Object size        & 52.1 & 54.5 & $-2.4$ & 55\% \\
Room size          & 41.4 & 41.4 & $\pm0.0$ & 100\% \\
Relative distance  & 36.1 & 32.5 & $+3.6$ & 8\% \\
Relative direction & 44.5 & 44.5 & $\pm0.0$ & 14\% \\
Route planning     & 32.5 & 31.4 & $+1.1$ & 20\% \\
Appearance order   & 41.1 & 28.5 & $+12.6$ & 14\% \\
\midrule
\textbf{Overall}   & 37.5 & 34.5 & $+3.0$ & 35\% \\
\bottomrule
\end{tabular}}
\end{table}

\subsection{Absolute Values Behind the Layer Ablation}
In Tab.~\ref{tab:ablation_main}, changes are reported so that a $0.000$ is translated as this component will not influence this metric. The absolute values it is based on are as follows. Those three variants which do not alter retrieval produce the same retrieval columns, and thus the text reports the differences.

\begin{table}[h]
\centering
\caption{Absolute suite scores for every ablation variant ($n{=}573$ relation,
$n{=}471$ retrieval).}
\label{tab:supp-abl-abs}
\scriptsize
\setlength{\tabcolsep}{2.5pt}
\resizebox{\columnwidth}{!}{
\begin{tabular}{lccccc}
\toprule
\textbf{Variant} & \textbf{Ver.\ all} & \textbf{Ver.\ answ.} & \textbf{SR@$0.5$} & \textbf{SR@$1.0$} & $\mathrm{Acc}_{\text{parent}}$ \\
\midrule
Full, RGB memory (reference) & 0.433 & 0.611 & 0.176 & 0.236 & 0.149 \\
Full, matched-input memory   & 0.553 & 0.662 & 0.263 & 0.359 & 0.242 \\
\midrule
w/o open-vocabulary object nodes & 0.079 & 0.616 & 0.028 & 0.042 & 0.034 \\
w/o wall anchors                 & 0.433 & 0.611 & 0.176 & 0.236 & 0.019 \\
w/o structural layer (flat)      & 0.433 & 0.611 & 0.176 & 0.236 & 0.000 \\
w/o relational hint in retrieval & 0.433 & 0.611 & 0.174 & 0.238 & 0.157 \\
\bottomrule
\end{tabular}}
\end{table}

\subsection{Navigation Metrics, Retained for Reference}
\label{sec:supp-nav-retained}
\begin{figure*}[t]
    \centering
    \includegraphics[width=\linewidth]{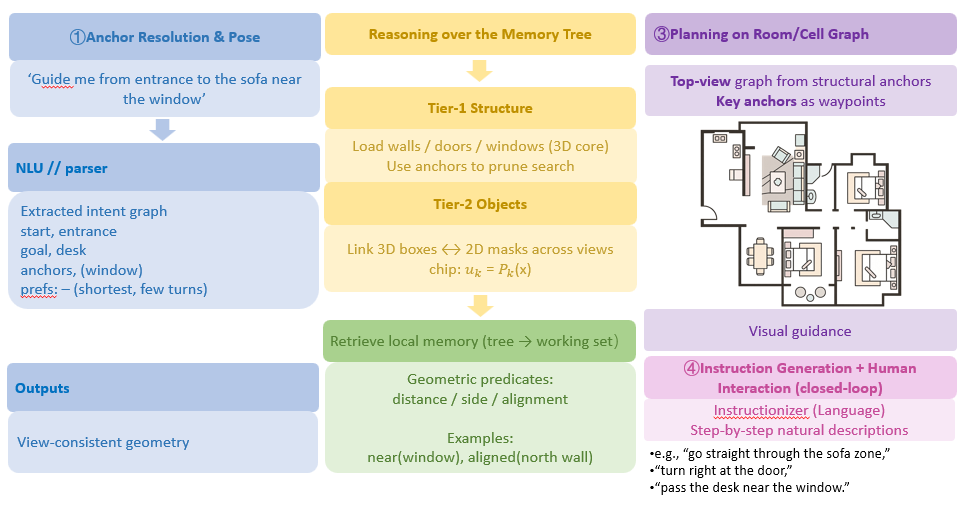}
    \caption{Step-by-step walkthrough of the navigation-style guidance demonstration: a command is parsed into an intent graph, resolved over the memory tree, planned with the help of an anchor-derived room/cell graph, and expressed as step-wise guidance. The given task is qualitative only; no quantitative statement uses these metrics.}
    \label{fig:supp_navigation}
\end{figure*}

The definitions below are retained for reference only; guidance is reported qualitatively and no number in the paper uses these metrics.

\subsubsection*{Task 2 --- Application: navigation-style guidance}

We have presented four measures of guidance: success, step completion, efficiency, and errors. The task is assessed at predetermined decision points (including but not limited to corners, turns and forks, and forward/branching decisions) instead of as a continuous low-level path planning.

\noindent\textbf{Success rate ($\mathrm{SR}_{\text{nav}}$).} An episode is successful if the final predicted goal lies within a threshold $d_{\mathrm{thr}}$ of the true target on the shared reference map; $\mathrm{SR}_{\text{nav}}$ is the fraction of successful episodes.

\noindent\textbf{Step completion ($\overline{\mathrm{StepComp}}$).} For episodes with sub-steps, we compute a weighted completion score per episode and then average:
\[
    \mathrm{StepComp}_i=\frac{\sum_{t} w_{i,t} c_{i,t}}{\sum_{t} w_{i,t}},
\]
\[
\overline{\mathrm{StepComp}}=\frac{1}{|\mathcal{D}|}\sum_i \mathrm{StepComp}_i.
\]

$c_{i,t}\in\{0,1\}$ shows whether the guidance at decision point t is properly completed and $w_{i,t}$ is its weight.

\noindent\textbf{Efficiency (SPL).} We use the standard Success-weighted by Path Length:
\[
\mathrm{SPL}=\frac{1}{|\mathcal{D}|}\sum_i S_i^{\text{nav}}\cdot \frac{L_i^\star}{\max(L_i,\,L_i^\star)},
\]
where $S_i^{\text{nav}}\!=\!1$ in case of successful episodes and $0$ otherwise. Both $L_i^\star$ and $L_i$ are plotted on the identical 2D floor-projected decision-point map, where $L_i^\star$ represents the shortest possible path to the goal and $L_i$ represents the actual guidance path due to the forecasted step sequence.

\noindent\textbf{Navigation error (NE).} Mean final distance to the goal on the same reference map, computed over failed episodes only. Distance thresholds, angular tolerances, and tie-breaking rules are specified in the Appendix.

\begin{figure*}[t]
    \centering
    \includegraphics[width=\linewidth]{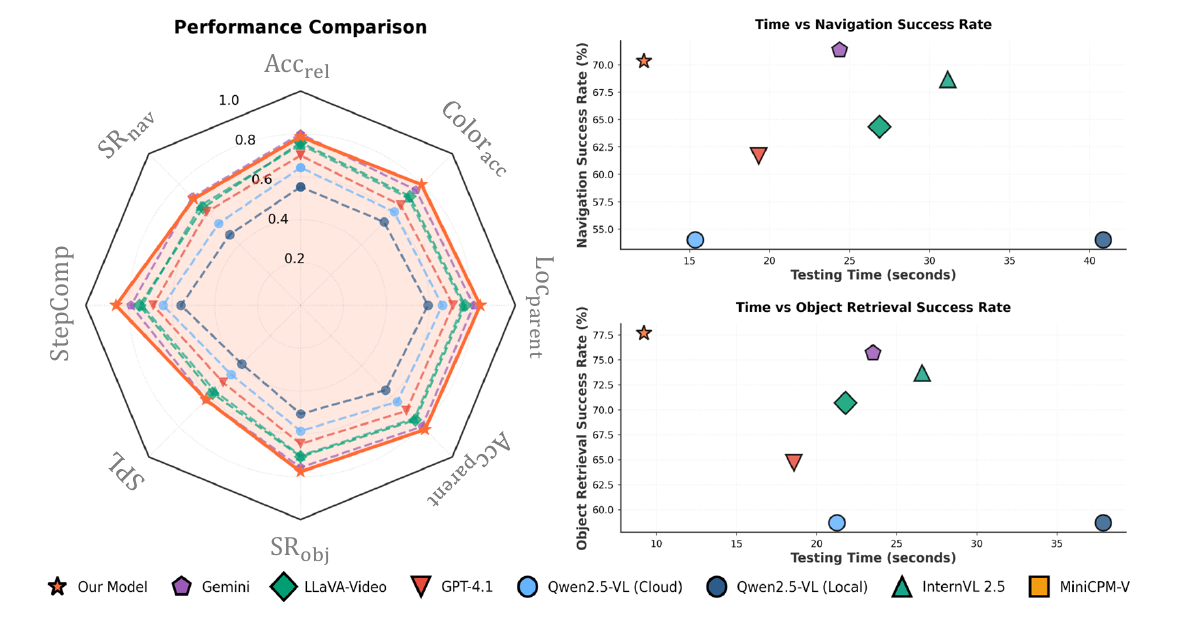}
    \caption{Summary on the three egocentric/Replica captures (not the public benchmarks). Left: radar over relative-position reasoning and guidance; each axis runs from $0$ at the centre to $1$ at the rim. Right: query time vs.\ success rate. Since these ground-truth episodes are suggested by the system and then human-checked, the figure is merely illustrative and does not support any head-to-head claim; all quantitative claims are in Tables $1$--$3$ of the main paper.}
    \label{fig:teaser_supp}
\end{figure*}

\section{Additional Diagnostics}
The two accountings as stated in the main text in summary form are presented below in full, without any changes made.

\subsection{Description Layers and Retrieval}
\noindent\textbf{Why we do not report a description-layer effect on retrieval.} Since the two description layers were intended to aid in answering a referential question, the obvious experiment is to remove them during retrieval. We did so and explain here why the measure is not available: descriptions are produced only for objects that are proposed by the layout stage, and they cover $34$ of the $471$ retrieval targets ($7\%$) in the RGB memories and none in the matched-input memories, which have their object layer supplied by the detection sweep. With $361$ of the $471$ queries carrying a referential cue, ablating the $34$ described targets cannot separate the layers at any meaningful spacing, and we prefer not to report a figure that noisy. What the layers are meant to deliver, resolving a hint against stored anchor-relative context, can be measured geometrically, and there it adds +0.002. With the colour measurement in this document, where the consolidation across views does not better the raw per-view captions, we do not state that the description layers enhance retrieval, and we have re-scoped the contribution appropriately: they attach attributes to a node, and it is the anchors that render the position of the node answerable.

\subsection{Candidate Selection versus Localisation}
On instance localisation the baseline remains clearly ahead ($0.41$ versus $0.26$). To see if the gap is the choice of candidates: an oracle permitted to choose the best among our label-matched candidates would achieve 0.48, whereas a deterministic selector that resolves the relational clue in the question (``the chair near the window'') against the memory recovers almost none of that headroom (+$0.002$). The oracle margin thus mostly contains objects that coincidentally lie close to the target rather than a disambiguation policy we are failing to apply, and the actual cost is localisation accuracy, box extent plus, for the RGB-only row, the registration residual. 

\subsection{Registration Residuals}
\noindent\textbf{The RGB-only retrieval number measures registration, not memory.} The scoring-time alignment has a median residual of $0.44$\,m against a $0.5$\,m success radius, and it fails outright on some scenes (residuals range from $0.08$\,m to $5.1$\,m). Separating them is informative: on the eight scenes where registration converges (residual $\le 0.6$\,m) the RGB-only memory is at $\mathrm{SR}_{\text{obj}}=0.211$ $[0.169,0.252]$ compared to $0.108$ $[0.057,0.158]$ on the four that fail, so the gap to the matched-input variant ($0.263$) narrows from $0.087$ to $0.052$ but does not disappear: alignment explains about $40\%$ of the RGB-only deficit and the estimated geometry itself explains the remainder. Increasing the radius to $1.0$\,m raises the full RGB-only figure from $0.176$ to 0.236. The fitted scale is centred a little above unity (median 1.11, interquartile range $1.01$--$1.23$, full range $0.85$--$1.37$ across the $12$ scenes), which bounds how much of the residual is a metric-scale error as opposed to a rotation or translation error. We take the RGB-only retrieval row as a lower bound with a quantified alignment component, and we regard the RGB-D row as the measurement that isolates the representation.

\section{Corpora, Ground Truth and Cost}
\label{sec:supp-corpora}
Tab.~\ref{tab:corpora} takes into account all the assessments of the main paper: provenance, scale, ground-truth origin, metrics, and one-time construction cost.

\begin{table*}[!t]
\centering
\caption{Each assessment in this paper: provenance, scale, ground truth, metrics, and construction cost. \emph{Independent} marks ground truth that does not originate in any system's reconstruction. Construction is done only once and is offline; the index column is the per-scene footprint a query touches, and each quantitative claim in Sec.~5 of the main paper is calculated using a metric listed here. The $12$ suite scene identifiers and the per-corpus question lists are available in the Supplementary Material.}
\label{tab:corpora}
\scriptsize
\setlength{\tabcolsep}{3pt}
\resizebox{\textwidth}{!}{
\begin{tabular}{llccllccll}
\toprule
& \multicolumn{3}{c}{\textbf{Provenance}} & \multicolumn{4}{c}{\textbf{Ground truth}} & \multicolumn{2}{c}{\textbf{Cost}} \\
\cmidrule(lr){2-4}\cmidrule(lr){5-8}\cmidrule(lr){9-10}
\textbf{Evaluation} & Scene source & \makecell{Scenes\\built} & \makecell{Questions /\\episodes}
& Origin & \makecell{Chance /\\prior} & \makecell{Indep.\\of recon.} & \makecell{Human\\verified}
& Metrics reported & \makecell{Build / scene\\(median)} \\
\midrule
\textbf{VSI-Bench}~\cite{vsibench} (Tab.~\ref{tab:vsi})
& ScanNet, ScanNet++, ARKitScenes & $287/288$ & $5{,}130$
& benchmark's released answers & $35.5$ freq.\ prior & \cmark & benchmark
& $8$ task scores, macro mean & $118$\,s \\
\textbf{OpenEQA}~\cite{openeqa} EM-EQA (Tab.~\ref{tab:openeqa})
& ScanNet portion & $79/89$ & $953$
& human-written answers & $33.5$ blind LLM & \cmark & benchmark
& LLM-Match, $7$ categories & $118$\,s \\
\textbf{ScanNet suite} (Tab.~\ref{tab:suite}), \emph{released}
& ScanNet, $12$ scenes & $12/12$ & \makecell[l]{$573$ relation, $471$ retrieval,\\$387$ description}
& \makecell[l]{official \texttt{aggregation.json}\\+ released instance boxes} & $0.500$ balanced & \cmark & \makecell[l]{$504$ episodes,\\$98.6\%$ confirmed}
& \makecell[l]{Ver.\ all / matched, $\mathrm{SR}_{\text{obj}}$@$0.5$/$1.0$,\\$\mathrm{Acc}_{\text{parent}}$, Colour} & \makecell[l]{$118$\,s RGB,\\$89$\,s RGB-D} \\
\midrule
Real egocentric probe set (supp., qualitative), \emph{released}
& \makecell[l]{$2$ self-captured + Replica} & $3/3$ & $174$ verified
& system-proposed, human-verified & --- & \xmark & $174$ episodes
& \makecell[l]{$\mathrm{Acc}_{\text{rel}}$, $\mathrm{Acc}_{\text{path}}$, Loc@parent/path,\\guidance metrics (qualitative)} & --- \\
\bottomrule
\end{tabular}}
\end{table*}

\section{Capability Matrix of Persistent-Memory Systems}
\label{sec:supp-capability}
Tab.~\ref{tab:capability} lists the positioning statement of the related-work section of the main paper: what every persistent-memory system keeps, and what should be provided to store them.

\begin{table*}[!t]
\centering
\caption{What persistent-memory systems store, and what they must be given to store it. \cmark{} present, \xmark{} absent, \pmark{} partial. \emph{Poses given} marks methods taking as input ground-truth camera poses, depth and intrinsics. \emph{Metric} marks a memory in which the distances and extents can be retrieved in metres. \emph{Anchor-first} marks a hierarchy based on structural features (walls, doors, windows) instead of on segmented regions. \emph{Train-free} marks methods which do not involve task-specific 3D or spatial training on top of off-the-shelf perception and language models. \emph{Index size} refers to the per-scene footprint that a query needs to hit.}
\label{tab:capability}
\scriptsize
\setlength{\tabcolsep}{3.5pt}
\resizebox{\textwidth}{!}{
\begin{tabular}{llcccccccll}
\toprule
& \multicolumn{2}{c}{\textbf{Input}} & \multicolumn{5}{c}{\textbf{Representation}} & & & \\
\cmidrule(lr){2-3}\cmidrule(lr){4-8}
\textbf{Method} & Sensor & Poses given
& \makecell{Persistent\\3D} & Metric & \makecell{Anchor-\\first} & \makecell{Open-vocab\\objects} & \makecell{Language\\descr.}
& \makecell{Train-\\free} & Index / scene & Evaluated on \\
\midrule
\multicolumn{11}{l}{\emph{Frame-sampling and instruction-tuned VLMs}} \\
GPT-4o, Gemini, Qwen2.5-VL & RGB video & \xmark & \xmark & \xmark & \xmark & \cmark & \cmark & \cmark & re-reads video & VSI-Bench, OpenEQA \\
VLM-3R~\cite{vlm3r} & RGB video & \xmark & \pmark\,implicit & \pmark\,implicit & \xmark & \cmark & \cmark & \xmark & in weights & VSI-Bench \\
\midrule
\multicolumn{11}{l}{\emph{Open-vocabulary 3D maps}} \\
OpenScene~\cite{openscene} & posed RGB-D + cloud & \req & \cmark & \cmark & \xmark & \cmark & \xmark & \xmark & dense features & ScanNet, Matterport3D \\
OpenMask3D~\cite{openmask3d} & posed RGB-D + cloud & \req & \cmark & \cmark & \xmark & \cmark & \xmark & \cmark & masks + features & ScanNet200, Replica \\
\midrule
\multicolumn{11}{l}{\emph{3D scene graphs}} \\
Hydra~\cite{hughes2022hydra} & RGB-D + odometry & \req & \cmark & \cmark & \pmark\,places/rooms & \xmark & \xmark & \cmark & --- & uHumans2, real robot \\
ConceptGraphs~\cite{conceptgraphs} & posed RGB-D & \req & \cmark & \cmark & \xmark\,bg.\ excluded & \cmark & \cmark & \cmark & --- & Replica, ScanNet, robot \\
HOV-SG~\cite{hovsg} & posed RGB-D & \req & \cmark & \cmark & \pmark\,floors/rooms & \cmark & \cmark & \cmark & --- & HM3DSem, ScanNet, Replica \\
GraphPad~\cite{graphpad} & posed RGB-D keyframes & \req & \cmark & \cmark & \pmark\,single room & \cmark & \cmark & \cmark & --- & OpenEQA \\
\midrule
\multicolumn{11}{l}{\emph{Persistent memories for embodied QA}} \\
3D-Mem~\cite{threedmem} & posed RGB-D & \req & \xmark\,snapshots & \xmark & \xmark & \cmark & \cmark & \cmark & snapshot images & OpenEQA, GOAT-Bench \\
VL-KnG~\cite{vlkng} & RGB video & \xmark & \xmark\,no 3D recon. & \xmark & \xmark & \cmark & \cmark & \cmark & --- & OpenEQA, NaVQA \\
\midrule
\textbf{SpatialMem (Ours)} & \textbf{RGB video} & \xmark & \cmark & \cmark & \cmark & \cmark & \cmark & \cmark & \textbf{14\,KB} & VSI-Bench, OpenEQA, ScanNet suite \\
\bottomrule
\end{tabular}}
\end{table*}

\section{OpenEQA: Protocol, Prompts and the Full Grid}
\label{sec:supp-openeqa}
\noindent\textbf{Scope.} The episodic-memory (EM-EQA) task, ScanNet portion: $953$ of its total $1{,}079$ questions over $79$ of $89$ scenes; the $10$ excluded scenes (all \texttt{scene07xx}) belong to the held-out ScanNet test split and have no publicly available data. The judge and metric are the benchmark's own: the LLM-Match prompt verbatim, a $1$--$5$ mark $r$ per answer, $s=(r-1)/4$, scaled to $0$--$100$; overall is the average over questions, the benchmark's convention. Judge model GPT-4o for every row; answering model GPT-4o for every row we execute (the published block uses the backbones its sources report, which is why it is context only).

\noindent\textbf{Rows.} The blind control is the benchmark's own language-prior prompt, verbatim. Serialised text renders the queryable index, room extent, structural layer, object nodes with metric positions and sizes, descriptions, as text, anchor-first, and answers from that text alone (this is the interface of the published caption-serialised rows). Uniform frames is the benchmark's frame prompt over $K$ frames chosen uniformly from the same frames the memory was constructed upon, hence each comparison is same-pixels-different-representation. Evidence-frame retrieval compares the noun phrases of the question with the object labels, votes on the per-object evidence-frame lists saved in the memory, attaches the top $K{=}8$ frames (uniform padding when fewer match), and uses the benchmark's frame prompt, the memory selects the pixels, the pixels respond. All runs are disk-cached and resumable; extending a row from the first $26$ built scenes to all $79$ reused every cached answer.

\noindent\textbf{The $2{\times}2$ on the matched subset} ($320$ questions on the $26$ scenes built first, identical questions throughout) separates frame selection from text serialisation: uniform-8 without text 61.6, evidence-8 without text 62.5, evidence-8 with the attached serialised text 57.6, text alone 31.5, blind 46.0. Selection is neutral-to-positive; the cost of attaching the serialisation is $4.9$ points, even when using the same images. At full scale ($953$) the final rows are as follows: blind 44.5, text 35.1, text without object nodes 29.0, uniform-8 61.6, evidence-8 $62.0$ $[59.7,64.3]$, GPT-4o@50 70.9, Gemini-2.5-Flash@50 72.1, Qwen2.5-VL-72B@50 63.1. Two additional text-mode ablations measured on the matched subset: without descriptions 32.0, Layer-1 only 33.1, descriptions do not significantly alter the text mode in either direction.

\noindent\textbf{Reading.} Anything that beats blind hands the answerer pixels. The memory's serialisation suppresses a strong model's priors ($35.1$ versus blind $44.5$), reproducing the published caption-serialised pattern (ConceptGraphs captions $34.4$ versus blind $35.5$); its retrieval is equivalent to uniform sampling in single-room captures, where eight uniform frames already cover the room, and beats blind precisely on the memory categories (localisation $48.0$ versus 34.8, spatial $55.4$ versus $41.2$). The natural regime for selective retrieval, long multi-room captures, is future work.

\section{Localisation Error and R@3: Method}
\label{sec:supp-mle}
mLE is the median of the distance between the returned instance centre and the annotated centre across the episodes a system answers ($471$/$431$/$392$ for ConceptGraphs, matched-input and RGB-only respectively; the baseline never abstains, hence its mLE conditions on nothing). Medians with $95\%$ bootstrap intervals: ConceptGraphs $1.03$ $[0.73,1.22]$, SpatialMem-RGBD $1.74$ $[1.59,1.93]$, SpatialMem $2.14$ $[1.91,2.32]$, sorted precisely by input privilege. R@3 asks whether a correct instance is among the top three candidates under the answerer's own hint scoring, expanded from argmin to a sort (the sole consistent extension); the recomputed top-1 reproduces the stored prediction on $471$/$471$ episodes for both rows, which is the trust gate the number has to pass before we report it. R@3@$0.5$\,m: matched-input $0.399$ $[0.36,0.44]$ against top-1 $0.263$ and label-matched oracle $0.479$; RGB-only $0.217$ $[0.18,0.26]$ against top-1 0.176. The answering code of the baseline records a single prediction, and therefore R@3 is undefined for it. Episode keys (scene $|$ ret $|$ target $|$ kind) are not unique, the same target may carry two queries with different hints, thus joins between the prediction files and the episode list are positional, never by key.

% ---- generated by suite/gen_ci_tables.py, do not hand-edit ----
\begin{table*}[t]
\centering
\caption{Bootstrap $95\%$ confidence intervals per task ($1{,}000$ resamples, seed $42$) of the three rows in Tab.~\ref{tab:vsi} that we ran ourselves. Resampling is within task, so the overall interval propagates each task's uncertainty through the macro mean.}
\label{tab:supp-vsi-ci}
\scriptsize
\setlength{\tabcolsep}{4pt}
\begin{tabular}{lccc}
\toprule
\textbf{Task} & \textbf{SpatialMem} & \textbf{w/o open-vocab.\ object nodes} & \textbf{LLaVA-NeXT-Video-7B, re-run} \\
\midrule
Object count & $24.9$\,{\tiny$[21.9,27.9]$} & $13.9$\,{\tiny$[11.5,16.4]$} & $48.6$\,{\tiny$[45.2,52.0]$} \\
Absolute distance & $27.5$\,{\tiny$[25.1,29.8]$} & $28.9$\,{\tiny$[26.4,31.2]$} & $13.9$\,{\tiny$[12.1,15.7]$} \\
Object size & $52.1$\,{\tiny$[49.8,54.4]$} & $54.5$\,{\tiny$[52.4,57.0]$} & $47.6$\,{\tiny$[45.4,49.7]$} \\
Room size & $41.4$\,{\tiny$[37.4,45.3]$} & $41.4$\,{\tiny$[37.4,45.6]$} & $24.4$\,{\tiny$[20.5,28.4]$} \\
Relative distance & $36.1$\,{\tiny$[32.2,39.4]$} & $32.5$\,{\tiny$[29.4,36.2]$} & $43.9$\,{\tiny$[40.0,47.8]$} \\
Relative direction & $44.5$\,{\tiny$[41.4,47.8]$} & $44.5$\,{\tiny$[41.4,47.6]$} & $42.6$\,{\tiny$[39.3,45.8]$} \\
Route planning & $32.5$\,{\tiny$[26.3,39.7]$} & $31.4$\,{\tiny$[24.7,38.1]$} & $33.5$\,{\tiny$[27.3,40.2]$} \\
Appearance order & $41.1$\,{\tiny$[37.2,45.1]$} & $28.5$\,{\tiny$[25.1,32.2]$} & $31.1$\,{\tiny$[27.0,34.6]$} \\
\midrule
\textbf{Overall} & $37.5$\,{\tiny$[36.1,38.8]$} & $34.5$\,{\tiny$[33.1,35.7]$} & $35.7$\,{\tiny$[34.2,37.0]$} \\
\bottomrule
\end{tabular}
\end{table*}

\begin{table*}[t]
\centering
\caption{The confidence intervals of Tab.~\ref{tab:suite} (n=$573$ relation, n=$471$ retrieval episodes after the human audit; $95\%$ bootstrap, $1{,}000$ resamples, seed $42$). The two relation columns have a common denominator; the answerable column eliminates the episodes where the queried entity cannot be found in the memory, hence it is only defined on our two rows and is equal to the all-episode value in systems that do not have the ability to abstain.}
\label{tab:supp-suite-ci}
\scriptsize
\setlength{\tabcolsep}{3pt}
\begin{tabular}{lcccc}
\toprule
\textbf{System} & \textbf{Relation (all)} & \textbf{Relation (answerable)} & $\mathrm{SR}_{\text{obj}}$ & $\mathrm{Acc}_{\text{parent}}$ \\
\midrule
ConceptGraphs & $0.611$\,{\tiny$[0.572,0.651]$} & $0.611$\,{\tiny$[0.572,0.651]$} & $0.412$\,{\tiny$[0.369,0.456]$} & $0.011$\,{\tiny$[0.002,0.021]$} \\
SpatialMem-RGBD & $0.553$\,{\tiny$[0.510,0.595]$} & $0.662$\,{\tiny$[0.620,0.706]$} & $0.263$\,{\tiny$[0.223,0.306]$} & $0.242$\,{\tiny$[0.206,0.285]$} \\
SpatialMem & $0.433$\,{\tiny$[0.393,0.473]$} & $0.611$\,{\tiny$[0.564,0.658]$} & $0.176$\,{\tiny$[0.144,0.214]$} & $0.149$\,{\tiny$[0.119,0.183]$} \\
GPT-4o & $0.613$\,{\tiny$[0.572,0.653]$} & $0.613$\,{\tiny$[0.572,0.653]$} & -- & -- \\
Gemini-2.5-Flash & $0.604$\,{\tiny$[0.564,0.646]$} & $0.604$\,{\tiny$[0.564,0.646]$} & -- & -- \\
Qwen2.5-VL-72B & $0.586$\,{\tiny$[0.545,0.628]$} & $0.586$\,{\tiny$[0.545,0.628]$} & -- & -- \\
GPT-4o-mini & $0.548$\,{\tiny$[0.506,0.586]$} & $0.548$\,{\tiny$[0.506,0.586]$} & -- & -- \\
\bottomrule
\end{tabular}
\end{table*}

\section{Scale Perturbation with the Registration Frozen}
\label{sec:supp-scalenoreg}
The $\pm10\%$ rows of the main ablation table re-apply the scoring-time similarity registration after perturbing the memory, as the reference run does, and they demonstrate almost no change, as expected, because a similarity fit absorbs a global scale error by design. Freezing the fit instead (fitting on the unperturbed memory, scoring the perturbed one) bounds how much the fit quietly repairs: SR@0.5 decreases by $0.036$ at $-10\%$ and $0.027$ at $+$10$\%$, with relation verification unchanged ($-0.004$/$-0.007$, relations are largely scale-invariant) and parent accuracy moving +$0.012$/$-0.017$. Files: \texttt{ours\_scale\_\{m10,p10\}\{,\_noreg\}\_pred.json}.

\section{Suite Columns Moved from the Main Table}
\label{sec:supp-suitecols}
\noindent\textbf{Per-predicate relation verification} (all episodes): each system is much stronger on the vertical family than on proximity, GPT-4o $0.79$/0.52, Gemini-2.5-Flash $0.76$/0.52, Qwen2.5-VL-72B $0.81$/0.47, GPT-4o-mini $0.82$/0.41, ConceptGraphs $0.85$/0.49, SpatialMem-RGBD $0.63$/0.51, SpatialMem $0.48$/$0.41$ (vertical/proximity, n=$197$/$376$). Our weakness relative to the scene graph sits in the vertical family; proximity is a draw. Hierarchy by annotated parent type, Door (n=$24$) and Window (n=$11$) columns: ConceptGraphs $0.21$/0.00, SpatialMem-RGBD $0.12$/0.18, SpatialMem $0.00$/0.00, sample sizes too small to separate the systems, which is why the main table keeps only Wall (n=$436$) and All.

\section{Probe-Set Data Card (Casual Egocentric Captures)}
\label{sec:supp-datacard}
\noindent\textbf{What is released.} Three indoor scenes, two casually recorded egocentric videos (a hotel room and a multi-room apartment, handheld phone capture, walking pace, motion blur and clutter present) and one public Replica room, along with the constructed memories, the $174$ human-verified evaluation episodes, the query-generation and verification protocol, and the evaluation code. Raw videos are published upon publication. The full dataset will be released under an open-source licence; the name of the dataset and the link to its hosting are withheld in this document to maintain the double-blind review process and will be given in the camera-ready version.

\noindent\textbf{What this set is and is not.} It is a probe set, as its goal is to cover a capture regime which does not manifest in public scanning datasets (handheld egocentric movement, blur, uncurated clutter), where we see frame-based models drop to near-chance relation accuracy whereas the memory's accuracy is preserved (Sec.~\ref{sec:supp-egocentric}). It is not a benchmark, because its ground truth is proposed by the system and then validated by humans, so it cannot include an episode that the system does not propose, and it must not be used to rank systems. The annotation-derived $12$-scene suite, with independent ground truth, $98.6\%$ generator precision under human audit and balanced probes, is the released artifact intended for ranking.

\section{Confidence Intervals}
\label{sec:supp-ci}
There are two main-text tables that quote only point estimates, for width. Here are their intervals, produced from the same prediction files the tables come from.

\section{Description Quality on the Audited Colour Labels}
\label{sec:supp-desc-audit}

The human audit supplied colour attributes for $175$ suite objects, which lets us measure the description layers on a larger sample than the five objects reported for the egocentric captures. Out of the $175$ objects, $43$ have a matching per-object description directory in our memory and $19$ contain usable per-view captions, an immediate consequence of the object-coverage limitation mentioned in the main text, because the audited objects are sampled from the official annotations and contain numerous small objects that our detector never proposes.

\begin{table}[h]
\centering
\scriptsize
\setlength{\tabcolsep}{3pt}
\begin{tabular}{lcc}
\toprule
\textbf{System} & \textbf{described objs.} & \textbf{all audited} \\
 & ($n{=}19$) & ($n{=}175$) \\
\midrule
Ours, Layer-1 (per-view) & 0.68 [0.47,0.90] & -- \\
Ours, Layer-2 (consolidated) & 0.42 [0.21,0.63] & -- \\
GPT-4o                                 & 0.74 [0.53,0.90] & 0.56 [0.49,0.63] \\
Gemini-2.5-Flash                       & 0.79 [0.58,0.95] & 0.59 [0.52,0.66] \\
Qwen2.5-VL-72B                         & 0.74 [0.53,0.90] & 0.69 [0.63,0.76] \\
GPT-4o-mini                            & \textbf{0.84} [0.68,1.00] & 0.61 [0.54,0.68] \\
\bottomrule
\end{tabular}
\caption{Colour recall in comparison with human labels, $95\%$ bootstrap intervals. Unlike the five-object measurement on our egocentric captures, the frame-based models recover colour attributes better than our description layers on these scenes, and the cross-view consolidation does not improve on the raw per-view captions. We report this rather than relying on the smaller sample. The intervals are wide and overlapping at n=19, and the consolidation figure is obtained through a re-implementation of the consolidation prompt.}
\end{table}

The consequence for the paper's claims is stated in the main text: attribute recognition is not a strength of this system, and the description layer is justified by what it attaches attributes to, i.e., a node in the metric hierarchy, rather than by recognition accuracy.

\section{Casual Egocentric Captures (Self-Proposed Ground Truth)}
\label{sec:supp-egocentric}
\noindent Ground truth here is proposed from the reconstruction and then confirmed by a human, so these numbers are not comparable to the annotation-derived suite in the main text and are reported separately. The Replica room is a clean public scan and scores at ceiling.

\label{sec:supp-realworld}

\subsection{Relation Understanding}
On the casual egocentric video the picture is strikingly different from the ScanNet views. We report two fair quantities: verification accuracy on balanced yes/no probes over viewpoint-invariant predicates with hard negatives (chance $0.5$), and cross-view consistency, i.e., whether a model returns the same lateral relation from three disjoint temporal segments; no canonical answer is assumed, hence the metric cannot punish a baseline unjustly. All five frame-based models sit at chance on verification ($0.50$--$0.53$) and are largely inconsistent across views ($0.12$--$0.64$), whereas SpatialMem is correct ($\sim$$0.99$) and view-invariant by construction ($1.00$) (Table~\ref{tab:relation}). The same models detect object presence well ($\mathrm{SR}_{\text{obj}}=0.91$--1.00), so the failure is specific to metric, cross-view relations under motion blur and clutter, the regime a persistent metric frame is meant to cover.

\begin{table}[t]
\centering
\caption{Spatial understanding vs.\ five frontier VLMs (pooled over the two egocentric captures and the public Replica room; $95\%$ CI). Verify: balanced yes/no on viewpoint-invariant relations (chance $0.5$); Consist: cross-view consistency on lateral relations; SR$_{\text{obj}}$: object-presence detection; Color: colour-attribute recall. SpatialMem's relation accuracy is reconstruction-limited (answered from geometry); baselines answer from frames. ``-'': free-text answers not reliably parsed.}
\label{tab:relation}
\scriptsize
\setlength{\tabcolsep}{5pt}
\begin{tabular}{lcccc}
\toprule
\textbf{Model} & \textbf{Verify}$\uparrow$ & \textbf{Consist}$\uparrow$ & \textbf{SR}$_{\text{obj}}\uparrow$ & \textbf{Color}$\uparrow$ \\
\midrule
GPT-4o            & 0.50 & 0.12 & 1.00 & 0.13 \\
Gemini-2.5-Flash  & 0.50 & 0.16 & 0.95 & 0.37 \\
Qwen2.5-VL-72B    & 0.53 & 0.60 & 0.91 & 0.13 \\
GPT-4o-mini       & 0.53 & 0.44 & 0.95 & 0.27 \\
Qwen2-VL-72B      & 0.53 & 0.64 & --   & --   \\
\midrule
\textbf{SpatialMem (Ours)} & \textbf{0.99} & \textbf{1.00} & \textbf{1.00} & \textbf{0.80} \\
\bottomrule
\end{tabular}
\end{table}

\subsection{Retrieval, Hierarchy, and Description}
In these captures SpatialMem retrieves the queried object ($\mathrm{SR}_{\text{obj}}=1.00$) and places it correctly in the memory tree ($\mathrm{Acc}_{\text{path}}=0.82$, $[0.64,0.96]$), the remainder being same-label disambiguation between visually identical objects located on different walls. Frame-based models confirm presence but do not produce the structural path. On these two captures the two-layer consolidation recovers $0.80$ of the ground-truth colour attributes against $\le\!0.37$ for the strongest baseline (Table~\ref{tab:relation}), but this is five objects and it does not generalise: on the $19$ audited ScanNet objects that our memory describes, frame-based models do better in recovering colour than we do ($0.74$--$0.84$ vs.\ $0.68$ for our per-view layer) and the cross-view consolidation is poorer than the raw captions ($0.42$). We therefore do not claim attribute recognition as a strength. The task of the description layer is to bind attributes to a node in the metric hierarchy, so that a retrieved object arrives with its structural context; a frame-based model can name a colour but cannot attach it to an instance in a persistent map (supplement, Sec.~\ref{sec:supp-desc-audit}).

\end{document}